\documentclass[lettersize, journal, 10pt]{IEEEtran}
\IEEEoverridecommandlockouts
\usepackage{makecell}
\usepackage{float}
\usepackage{cite}
\usepackage{textcomp}
\usepackage{xfrac}
\usepackage{amsmath,amssymb,amsfonts}
\usepackage[linesnumbered,ruled,vlined]{algorithm2e}
\usepackage{tikz}
\usetikzlibrary{shapes,arrows}
\usepackage{graphicx}
\usepackage{subcaption}
\usepackage{colortbl}
\usepackage{textcomp}
\usepackage{multicol}
\usepackage{multirow}
\usepackage{siunitx}
\usepackage{hyperref}
\usepackage{orcidlink}
\usepackage{xcolor}
\def\BibTeX{{\rm B\kern-.05em{\sc i\kern-.025em b}\kern-.08em
    T\kern-.1667em\lower.7ex\hbox{E}\kern-.125emX}}

\usepackage{hyperref}
\hypersetup{colorlinks=true,linkcolor=blue,urlcolor=black,citecolor=blue}
\newcommand{\myhyperref}[2]{\hyperref[#1]{#2 \ref*{#1}}}

\makeatletter
\def\ps@IEEEtitlepagestyle{%
  \def\@oddfoot{\mycopyrightnotice}%
  \def\@oddhead{\hbox{}\@IEEEheaderstyle\leftmark\hfil\thepage}\relax
  \def\@evenhead{\@IEEEheaderstyle\thepage\hfil\leftmark\hbox{}}\relax
  \def\@evenfoot{}%
}
\def\mycopyrightnotice{%
  \begin{minipage}{\textwidth}
  \centering \scriptsize
  Copyright~\copyright~2024 IEEE. Personal use of this material is permitted. Permission from IEEE must be obtained for all other uses, in any current or future media, including\\reprinting/republishing this material for advertising or promotional purposes, creating new collective works, for resale or redistribution to servers or lists, or reuse of any copyrighted component of this work in other works by sending a request to pubs-permissions@ieee.org.
  \end{minipage}
}
\makeatother

\begin{document}

\title{R2 Indicator and Deep Reinforcement Learning Enhanced Adaptive Multi-Objective Evolutionary Algorithm}

\author{\IEEEauthorblockN{Farajollah Tahernezhad-Javazm \orcidlink{0000-0002-5073-9802},}
\and
\IEEEauthorblockN{Debbie Rankin \orcidlink{0000-0003-2110-0599},}
\and
\IEEEauthorblockN{Naomi Du Bois \orcidlink{0000-0002-9350-2100},}
\and
\IEEEauthorblockN{Alice E. Smith \orcidlink{0000-0001-8808-0663}, \textit{Life Fellow, IEEE,}}
\and
\IEEEauthorblockN{ and Damien Coyle \orcidlink{0000-0002-4739-1040}, \textit{Senior Member, IEEE}}
\thanks{This work has been submitted to the IEEE for possible publication. Copyright may be transferred without notice, after which this version may no longer be accessible.}
\thanks{Farajollah Tahernezhad-Javazm, Debbie Rankin, Naomi Du Bois, and Damien Coyle are with Intelligent Systems Research Center, School of Computing, Engineering and Intelligent Systems, Ulster University, Londonderry, BT48 7JL, UK. (email: tahernezhad\_javazm-f@ulster.ac.uk)}
\thanks{Damien Coyle and Naomi Du Bois are also with the Bath Institute for Augmented Human, University of Bath BA2 7AY, Bath UK.}
\thanks{Alice E. Smith is with the Department of Industrial and Systems Engineering and Department of Computer Science and Software Engineering, Auburn University, Auburn, AL 36849, USA.}}


\maketitle

\begin{abstract}
Choosing an appropriate optimization algorithm is essential to achieving success in optimization challenges. Here we present a new evolutionary algorithm structure that utilizes a reinforcement learning-based agent aimed at addressing these issues. The agent employs a double deep q-network to choose a specific evolutionary operator based on feedback it receives from the environment during optimization. The algorithm's structure contains five single-objective evolutionary algorithm operators. This single-objective structure is transformed into a multi-objective one using the R2 indicator. This indicator serves two purposes within our structure: first, it renders the algorithm multi-objective, and second, provides a means to evaluate each algorithm’s performance in each generation to facilitate constructing the reinforcement learning-based reward function. The proposed R2-reinforcement learning multi-objective evolutionary algorithm (R2-RLMOEA) is compared with six other multi-objective algorithms that are based on R2 indicators. These six algorithms include the operators used in R2-RLMOEA as well as an R2 indicator-based algorithm that randomly selects operators during optimization. We benchmark performance using the CEC09 functions, with performance measured by inverted generational distance and spacing. The R2-RLMOEA algorithm outperforms all other algorithms with strong statistical significance ($p<0.001$) when compared with the average spacing metric across all ten benchmarks. 
\end{abstract}


\section{Introduction}\label{sec1}

\IEEEPARstart{D}{ue} to the increasing complexity and difficulty of real-world problems, more reliable optimization techniques have become necessary in recent decades. There are numerous types of optimization problems, including single-objective/multi-objective (many-objective), continuous/discrete, and constrained/unconstrained. In recent literature, effort has been concentrated on multi-objective problems and many-objective problems (MOPs/MaOPs). MOPs/MaOPs can be addressed using two well-known strategies: mathematical-based approaches and evolutionary algorithms (EAs)/swarm-based intelligence (SI) methods. Although EAs and SI are more time-consuming compared to many mathematical-based methods, they are flexible to implement and more effective in non-differential, noisy, and complicated environments \cite{mukhopadhyay2013survey}.

Generally, multi-objective evolutionary algorithms (MOEAs) can be divided into three main groups when dealing with MOPs/MaOPs: domination-based, decomposition-based, and indicator-based methods. Indicator-based algorithms have garnered significant attention recently in the field of MOEAs. They evaluate and rate the performance of each individual in an MOEA using a set of criteria derived from the indicator (convergence and distribution). The first MOEA method to use the indicator concept to solve MOPs is the Indicator-Based Evolutionary Algorithm (IBEA) \cite{zitzler2004indicator}. IBEA, which applies hypervolume (HV), as the indicator, suffers from the problem of high computational cost, especially in MaOPs. To address this issue, many indicators were introduced in recent years, such as R2 \cite{brockhoff2012properties}, $e^{+}$ \cite{zitzler2004indicator}, $\Delta_p$ \cite{schutze2012using}, spacing (SP) \cite{schott1995fault}, and inverted generational distance (IGD) \cite{coello2004study}. Approaches based on dominance and decomposition are more prevalent than methods based on indicators; however, indicator-based techniques have recently demonstrated promising performance compared to other methods.

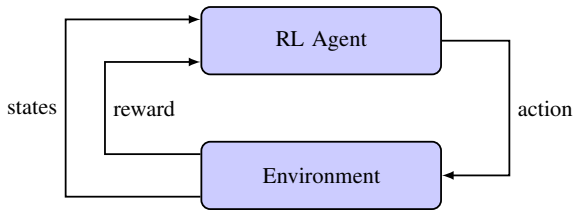
\begin{figure}[ht]
\center 
\tikzstyle{block} = [rectangle, draw, 
    text width=10em, text centered, rounded corners, minimum height=3em]
    
\tikzstyle{line} = [draw, -latex]
\scalebox{0.85}{
\begin{tikzpicture}[node distance = 6em, auto, thick]
    \node [block, fill = blue!20] (Agent) {RL Agent};
    \node [block,  fill = blue!20, below of=Agent] (Environment) {Environment};
    
     \path [line] (Agent.0) --++ (3em,0em) |- node [near start]{action} (Environment.0);
     \path [line] (Environment.190) --++ (-6em,0em) |- node [near start] {states} (Agent.170);
     \path [line] (Environment.170) --++ (-4.25em,0em) |- node [near start, right] {reward} (Agent.190);
\end{tikzpicture}}
\caption{General reinforcement learning block diagram} \label{fig.RL}
\end{figure}

Each EA possesses unique features in terms of the two exploration and exploitation criteria. Consequently, different EA methods result in different outputs. Selecting an appropriate EA for a specific problem is a time-consuming and tedious task, requiring a knowledgeable expert. In addition, during the optimization process, the dynamics of the problems change constantly, and applying specific operators/parameters is crucial for success. Tuning and controlling approaches are two techniques aimed at improving the selection of EA settings. The fundamental distinction between these methods lies in the timing of parameter/operator selection. Prior to the optimization process, EA settings are adjusted in parameter tuning. Alternatively, parameters/operators are tuned during the execution time in the controlling methods, also known as 'controlling on the fly' \cite{de2021systematic, karafotias2014parameter}. According to the nature of real-world problems, controlling methods\textemdash which can be divided into two categories: predictive methods and reinforcement-based methods\textemdash are better suited to tackling complex situations \cite{de2021systematic}. Predictive-based approaches employ statistical and machine learning models to foresee the best parameters/operators that will impact the future optimization process. In fact, these approaches use the past performance of EAs as a time series to forecast future parameters \cite{aleti2011predictive, aleti2014choosing}. Reinforcement Learning (RL) is an optimization technique that follows the principles of the Markov Decision Process (MDP) and operates based on the agent's interactions with its environment \cite{sutton2018reinforcement}. As illustrated in \myhyperref{fig.RL}{Figure}, the RL process comprises two main parts: agent and environment. It is a closed-loop process in which the agent performs an action based on feedback acquired from the environment (states and rewards), and the environment subsequently responds to the agent's action. Until the agent learns how to cope with the environment, this trial-and-error process continues. The RL agent is responsible for choosing the best action for maximizing the final accumulated reward. RL is an efficient nature-inspired method for modifying EAs, particularly when dealing with complex conditions \cite{drugan2019reinforcement}. In general, RL can be classified into two well-known main categories: temporal difference-based and policy gradient-based methods. Temporal difference-based methods can be utilized in problems with discrete action space (operator selection), while policy gradient-based methods are suitable for continuous action space problems (parameter selection). Many researchers have addressed the combination of EAs with RL; however, the majority concentrated on single-objective evolutionary algorithms (SOEAs) such as the genetic algorithm (GA) \cite{pettinger2002controlling, chen2005scga, eiben2006reinforcement} or differential evolution (DE) \cite{sallam2020evolutionary, ning2018reinforcement, visutarrom2020reinforcement, tan2022differential} algorithms.

Considering the aforementioned issues, we propose an adaptive MOEA structure that can deal with MOPs and additionally is not limited to a single or specific EA. As the dynamics of MOPs vary with each generation of optimization, MOEAs require different operators to handle these dynamics, dependent on the criteria of exploration and exploitation. Five EAs were therefore deployed in our work to meet this need. We transformed five single-objective EAs—GA \cite{holland1992adaptation}, evolutionary strategy (ES) \cite{beyer2002evolution}, teaching learning-based optimization (TLBO) \cite{rao2011teaching}, whale optimization algorithm (WOA) \cite{mirjalili2016whale}, and equilibrium optimizer (EO) \cite{faramarzi2020equilibrium}—into multi-objective EAs by using the R2 indicator. EO and WOA are two relatively new EAs known for their fast convergence \cite{al2023equilibrium, rana2020whale, tahernezhad2022r2}. In contrast, GA and ES are more established EAs. ES is particularly good at exploration \cite{hansen2001completely, hansen2015evolution}, while TLBO has shown to be effective at balancing the trade-off between exploration and exploitation \cite{zou2019survey, tahernezhad2022hybrid, tahernezhad2018review}. In our architecture, a double deep Q-learning Network (DDQN) \cite{mnih2015human} is used as the hyper meta-heuristic to select the appropriate EA in each generation based on the optimization process's feedback. The proposed algorithm, called R2-RLMOEA, is tested on the CEC09 multi-objective benchmarks, and the results are compared to the five MOEAs listed previously without guidance from the RL agent during the process. Based on the two indicators of IGD and SP, the results demonstrate the state-of-the-art performance of our proposed structure compared with the other algorithms.

The remainder of the paper is organized as follows: a literature review of recent works in the field of RL-EA is provided in \myhyperref{Sec.RelatedWorks}{Section}. Generic principles of multi-objective problems and Q-learning-based models are explained in \myhyperref{Sec.Background}{Section}. \myhyperref{Sec.OurAlg}{Section} introduces the combined EA-RL structure proposed in this work (R2-RLMOEA). Experimental settings and results are described in \myhyperref{Sec.Setting}{Section} and \myhyperref{Sec.Results}{Section}, respectively. The results are discussed in \myhyperref{Sec.Disc}{Section}. Finally, the conclusion is provided in \myhyperref{Sec.Conclusion}{Section}.

\section{Related works}\label{Sec.RelatedWorks}
Regarding the combination of RL with EA, DE has received the most attention as a well-known EA in recent years. The essence of DE is based on the principle of mixing population variety to develop superior individuals. Like many other EAs, it employs the three operators of selection, crossover, and mutation, in which the two specific parameters of crossover probability (Cr) and scale factor (F) play a crucial role.

Weikang et al. \cite{ning2018reinforcement} proposed a multi-objective evolutionary algorithm based on an RL controller and the multi-objective evolutionary algorithm based on decomposition (MOEA/D) \cite{zhang2007moea} with Tchebycheff as the scalarization function. However, compared to other works, they use the state-action-reward-state-action (SARSA) algorithm \cite{sutton2018reinforcement} instead of Q-learning to choose the number of neighbors and type of DE mutation operators during optimization. The proposed structure is evaluated with fitness-rate-rank-based multi-armed bandit (FRRMAB) \cite{li2013adaptive} using the two metrics of IGD and HV \cite{zitzler1999multiobjective} on ten benchmark functions from CEC09. In terms of the HV and IGD metrics, RL-MOEA/D performed better than FRRMAB on most of the test instances, except for UF4, UF5, UF6, and UF10. Although both algorithms achieved similar results, the median values obtained by RL-MOEA/D were better than those obtained by FRRMAB.

Tan et al. \cite{tan2022differential} introduced reinforcement learning-based hybrid parameters and mutation strategies differential evolution (RL-HPSDE) based on Q-learning to improve the DE optimization process through parameter tuning and operator selection on single-objective benchmark problems from CEC17. Four states are identified based on the two methods of dynamic fitness distance correlation (DFDC) \cite{malan2013survey} and dynamic ruggedness of information entropy (DRIE) \cite{huang2020fitness}. Also, actions are defined based on combining two distribution functions (Levy and Gauchy distribution) and two mutation operators (DE/current to rand/1 and DE/current to best/1). In addition, they applied a variant population policy (decreasing) in their work to improve the algorithm's performance. In this study, the performance of five algorithms - JADE, SHADE, LSHADE, iLSHADE \cite{brest2016shade}, and jSO \cite{brest2017single} - were compared across different dimensions (10, 30, 50, and 100). The results showed that RL-HPSDE outperformed the other algorithms in 10, 30, and 50 dimensions. However, in 100 dimensions, jSO performed better than RL-HPSDE. Based on the results, RL-HPSDE is ranked first, followed by jSO. iLSHADE and LSAHDE, which share a similar ranking, while JADE and SHADE are ranked last.

In addition to F and Cr, population size (N) can significantly affects DE performance. A big F or a small F lead to more exploration and exploitation, respectively. On the other hand, the best values of Cr and N vary with each particular problem \cite{price2008eliminating, qin2005self, ilonen2003differential}. Jianyong et al. \cite{sun2021learning} proposed a policy gradient-based RL (Learning Differential Evolution [LDE]) structure in continuous action space for tuning F and CR. They applied a Long Short-Term Memory (LSTM) network to consider the nature of MDP-based problems and the state dependencies. They utilized CEC13 (for training and testing) and CEC17 (for testing) benchmark functions to demonstrate the effectiveness of their framework. The algorithm was tested on benchmarks with 10 and 30 dimensions. Overall, it outperformed variations of conventional DE algorithms but was inferior to jSO and HSES \cite{zhang2018hybrid} in most of the benchmark tests, especially jSO. The drawback of their structure was the long training time, exacerbated by increasing the number of decision variables. LSTMs are significantly impacted by the number of neurons in hidden layers. They found that there is a complex relationship between the best number of neurons in hidden layers and each benchmark function.

Lue et al. \cite{tao2022differential} tried to address the real-world multi-objective assembly line feeding problem with an RL-based multi-objective DE. They applied a decomposition-based method for converting multi-objective to single-objective problems with an external archive for reserving non-dominated solutions in their framework. They applied three criteria of success rate, domination, and distribution to represent their states and to control convergence and distribution. Additionally, they utilized three mutation operators: DE/rand/2, DE/current-to-best/1, and DE/current-to-best/1. They determined that the proposed framework outperformed the three MOEAs of NAGA-II, NSGA-III \cite{deb2013evolutionary}, and SMPSO \cite{nebro2009smpso} in terms of convergence and solution quality.

Michele and Giovanni \cite{tessari2022reinforcement} proposed a generic model for parameter tuning in continuous action space. Proximal Policy Optimization (PPO) \cite{schulman2017proximal}, a policy gradient-based RL, is used as the agent for optimizing the parameters of two well-known EA algorithms: Co-variance Matrix Adaptation Evolutionary Strategies (CMA-ES) \cite{hansen2001completely} and DE. The agent chooses the proper step size (CMA-ES), scaling factor (DE), and crossover rate (DE) during the optimization process. Performance indicators include area under the curve, and best fitness of Run \cite{shala2020learning}. In addition, the reward function is considered based on the inter-generational $\Delta f$ to demonstrate optimization performance improvement and robustness across different objective scales. The results emphasize the importance of reward normalization and observation space design in achieving the best outcomes.

Many publications have explored the integration of RL with EAs for operator and parameter selection, as summarized in \myhyperref{tab.RecentRLEA}{Table}. It is evident that most of these works concentrate on SOEAs, and even those addressing MOEAs predominantly focus on integration with DE operators. R2-RLMOEA distinguishes itself by targeting MOEAs through the innovative use of the R2 indicator, enhancing the dynamic decision-making capabilities of RL by incorporating a diverse array of EAs for optimization. This approach not only broadens the applicability of RL-EA integration to complex multi-objective optimization tasks but also leverages the unique strengths of various evolutionary strategies to navigate the optimization landscape more effectively.

\begin{table}[ht]
  \caption{Publications in the field of RL-EA (operator/parameter selection)}
  \label{tab.RecentRLEA}
  \center
  \scalebox{0.7}{
  \begin{tabular}{|c|c|c|c|}
    \hline
    Reference & RL algorithm & Type & EA algorithm \\
    \hline
    \cite{sun2021learning} & Policy Gradient algorithm & SOEA & DE \\
    \hline
    \cite{tessari2022reinforcement} & PPO & SOEA & CMA-ES \& DE \\
    \hline
    \cite{de2019learning} & Dueling Double Deep Q-Learning & SOEA & PSO \cite{kennedy1995particle} \\
    \hline
    \begin{tabular}{@{}c@{}} \cite{chen2005scga} \\ \cite{sakurai2010method} \end{tabular} & SARSA & SOEA & GA \\
    \hline
    \cite{dantas2021online} & Deep Q-Network & SOEA & Hyper-Heuristic Algorithm \\
    \hline
    \cite{eiben2006reinforcement}   & Q-Learning \& SARSA & SOEA & GA \\
    \hline
    \cite{gambardella1994reinforcement}   & Q-Learning & SOEA & ACO \\
    \hline
    \cite{handoko2014reinforcement} & Q-Learning & SOEA & memetic search \\
    \hline
    \cite{karafotias2014comparing} & Q-learning & SOEA & \begin{tabular}{@{}c@{}} ES \\ Cellular GA \cite{alba2008introduction} \\ GA MPC \cite{elsayed2011ga} \\ IPO-10DDrCMAES \cite{liao2013bounding}\end{tabular}\\
    \hline
    \cite{emary2017experienced} & RL principle & SOEA & GWO \cite{mirjalili2014grey} \\
    \hline
    \cite{sadhu2018synergism} & Q-Learning & SOEA & Firefly Algorithm \cite{yang2010nature} \\
    \hline
    \cite{sharma2019deep} & Double Deep Q-learning & SOEA & DE \\
    \hline
    \begin{tabular}{@{}c@{}} \cite{tan2022differential} \\ \cite{sallam2020evolutionary} \end{tabular} & Q-learning & SOEA  & DE \\
    \hline
    \begin{tabular}{@{}c@{}} \cite{li2019differential} \\ \cite{tao2022differential} \end{tabular} & Q-learning & MOEA & DE \\
    \hline
    \cite{zou2021reinforcement} & Q-learning & MOEA & NSGA-II-DE \\
    \hline
    \cite{ning2018reinforcement} & SARSA & MOEA &  DE operators \& MOEA/D \\
    \hline
    R2-RLMOEA & Double Deep Q-learning & MOEA & \begin{tabular}{@{}c@{}} ES, GA, TLBO \\ WOA, EO, R2 Indicator \end{tabular} \\
    \hline
\end{tabular}}
\end{table}

\section{Preliminaries}\label{Sec.Background}
This section explains the algorithms and methods used in our paper. It includes the definition of multi-objective optimization and the concept of the R2 indicator for solving MOPs. Additionally, the fundamental building block of R2-RLMOEA, Double Deep Q-learning, is described.
\subsection{Multi-Objective Optimization Problem (MOP)}
Generally, a MOP such as  $\textbf{F}(\textbf{x})$ with the $"m"$ objective functions of ($f_1(\textbf{x}),..., f_m(\textbf{x})$) can be shown as \myhyperref{Eq.MOP}{Equation}:

\begin{align}\label{Eq.MOP}
  \mathrm{minimize} \:\: \textbf{F}(\textbf{x}) &= (f_1(\textbf{x}),..., f_m(\textbf{x}))^T \\
  \forall \textbf{x} &\in \Psi \: (\Psi^n \to R^m) \nonumber
\end{align}

\noindent where $\textbf{x}=(x_1, x_2,...,x_n)^T$ is a vector with dimension $n$ ($n$ is the number of decision variables), and $\Psi^n$ and $R^m$ are the search/decision space and objective space, respectively. The goal of \myhyperref{Eq.MOP}{Equation} is to minimize/maximize all objective functions ($f_1(x),..., f_m(x)$) in one run and find Pareto optimal solutions and Pareto Front (PF).

A vector $\textbf{a} = (a_1, a_2,..., a_m)^T$ dominates vector $\textbf{b} = (b_1, b_2,..., b_m)^T$, denoted by $\textbf{a} \prec \textbf{b}$, iff $\forall i \in \{1, 2, ..., m\}, f_i(\textbf{a})\le f_i(\textbf{b})$ and $f_k(\textbf{a}) \ne f_k(\textbf{b}), \forall k \in \{1, 2, ..., m\}$ . In addition, a vector in decision space such as $\textbf{x}^* \in \Psi$ is named a Pareto optimal solution if there is no $\textbf{y}\prec \textbf{x}^*$, iff $\forall \textbf{y}$ $\exists$ $\Psi$. The Pareto optimal set (PS) is the set of Pareto optimal solutions, and PF is the projection of PS from decision space into objective space ($ PF = \{\textbf{F}(\textbf{x}) \in R^m | \textbf{x} \in PS \}$).

\subsection{R2-Indicator}

Although HV is one of the most popularly employed indicators, the R2 indicator and its family (R1 and R3) \cite{hansen1994evaluating} is another indicator that evaluates both convergence and diversity \cite{falcon2020indicator}. Compared to HV, R2 requires substantially fewer computations and provides more distributed solutions \cite{hernandez2015improved}. The generic functionality of the R2 indicator is to evaluate the quality of two populations. To compute the R2 indicator, it is necessary to identify both the scalarization function and the reference point, and different scalarization functions have different effects on the results \cite{hansen1994evaluating}. We utilized the achievement scalarization function (ASF) as the utility function in our study. R2 can thus be calculated as \myhyperref{Eq.R2}{Equation} for the population/set of $\mathcal{P}$, the reference/utopian point $\textbf{z}^*$, and the weight vector $\textbf{w} = (w_1, w_2, ..., w_m) \in \Omega $  $(\parallel \textbf{w} \parallel_1 = 1 $ and $w_1, w_2, ..., w_m \geq m )$.

\begin{equation}\label{Eq.R2}
  R2(\mathcal{P}, \Omega, \textbf{z}^* ) = \frac{1}{|\Omega|} \sum_{\textbf{w} \in \Omega} \underset{\textbf{p} \in \mathcal{P}}{min} \left\{ASF(\textbf{p}|\textbf{w}, \textbf{z}^*)\right\}
\end{equation}

\noindent where ASF is the scalarization function and can be defined as \myhyperref{Eq.ASF}{Equation}: 

\begin{equation}\label{Eq.ASF}
ASF(\textbf{p}|\textbf{w}, \textbf{z}^*)= \underset {1 \leq i \leq m}{\max} \; \left\{\frac{\left|p_i-z^{*}_i\right|}{w_i}\right\}
\end{equation}

The optimal result is related to a smaller value of R2, which indicates a shorter distance between the reference point $\textbf{z}^*$ and the arbitrary set $\mathcal{P}$. The reference point is an ideal point to which no other point in the population can dominate.

\subsection{Double Deep Q-learning (DDQN)}
One of the most popular types of RL algorithm is Q-learning \cite{jang2019q}, which is a model-free and off-policy method and, like other types of RL, includes four parts: environment, state, action, and reward. The goal is to learn a policy enabling the agent to take optimal actions and maximize the rewards. In Q-learning, an agent makes decisions based on the action-valued function (derived from the Bellman equation), which can be defined as \myhyperref{Eq.QLearning}{Equation}. 

\begin{align}\label{Eq.QLearning}
  Q(s_t, a_t) &\leftarrow Q(s_t, a_t) + \alpha \left[ r_{t+1} + \gamma \; \underset{a \in \mathcal{A}}{\max} \right. \nonumber \\
  &\quad \left. Q(s_{t+1}, a) - Q(s_t, a_t) \right]
\end{align}

\noindent where $\mathcal{A} = [a_1, a_2, ...., a_n]$ is the set of possible actions. $a_t$ and $s_t$ denote action and state at time $t$, respectively. After performing action $a_t$, $r_{t+1}$ is the earned reward/punishment. $\alpha$ is known as the learning rate, specifying the acceleration of learning, and $\gamma$ is a value between zero to one. $\gamma = 0$ converts all the future rewards to zero. Thus, the agent is promoted by immediate rewards. On the other hand, if gamma is equal to one, we give the agent more patience to develop its long-term policy. \cite{sutton2018reinforcement}. Q-learning uses a look-up table (Q-table) to map various states to various actions and addresses problems with discrete states/actions. When the states are large or continuous, it cannot function properly. Mnih et al. \cite{mnih2013playing} attempted to tackle this issue by substituting a neural network for the Q-table. In deep Q-learning (DQN), the neural network acts as the function approximator for estimating the Q-value, leading to the selection of the subsequent action. Additionally, its performance is enhanced by introducing the concept of an experience replay buffer \cite{lin1992self}. Despite its success in many problems, DQN can suffer from action overestimation (resulting in premature convergence), unstable training, and poor performance. This problem is solved in double deep Q-learning (DDQN) \cite{van2016deep} by adding another neural network. In DDQN, the first/main network ($Q^{m}$) is used for action selection, while a second/target network ($Q^{t}$), updated periodically based on the main network, is used for action evaluation. 

\begin{align}\label{Eq.DDQLearning}
  Q^{m}(s_t, a_t; \theta) &\leftarrow Q^{m}(s_t, a_t; \theta) + \alpha \Big[ r_{t+1} + \gamma \; Q^{t}\Big(s_{t+1}, \nonumber \\
  &\quad \underset{a \in A}{\arg\max} \; Q^{m}(s_{t+1}, a; \theta); \theta^{'}\Big) - Q^{m}(s_t, a_t; \theta)  \Big]
\end{align}

\noindent where $\theta$ and $\theta^{'}$ are the main and target network parameters, respectively. Based on the DDQN (\myhyperref{Alg.DDQN}{Algorithm.}), removing the maximization of the next state leads to reducing bias maximizing.

\IncMargin{1em}
\begin{algorithm}[ht]
\SetKwData{Left}{left}\SetKwData{This}{this}\SetKwData{Up}{up}
\SetKwFunction{Union}{Union}\SetKwFunction{FindCompress}{FindCompress}
\SetKwInOut{Input}{input}\SetKwInOut{Output}{output}
\Input{Initialized $Q^{m}$ and $Q^{t}$, update limit ($\tau$), counter ($c = 0$) \\
maximum iteration ($\mathcal{N}_{game}$), empty experience replay buffer ($\mathcal{R}$)}
\Output{trained $Q^{m}$}
\BlankLine
\For{$n \in \mathcal{N}_{game}$}{
    Observe $s_t$ and choose $a_t \in \mathcal{A}$ \\
    apply $a_t$, go to $s_{t+1}$ and receive $r_{t+1}$ \\
    store $(s_t, a_t, r_{t+1}, s_{t+1})$ $\in$ $\mathcal{R}$ }
    \For{each update step}{
    $c \leftarrow c + 1$ \\
    sample a batch of $(s_t, a_t, r_{t+1}, s_{t+1})$ $\in$ $\mathcal{R}$ \\
    $\mathcal{P} = r_{t+1} + \gamma \; Q^{t}(s_{t+1}, \underset{a \in \mathcal{A}}{argmax} \; Q^{m}(s_{t+1}, a; \theta), \theta^{'}) - Q^{m}(s_t, a_t, \theta)$ \\
    $ Q^{m}(s_t, a_t) \leftarrow Q^{m}(s_t, a_t) + \alpha \; \mathcal{P}$ \\ 
    \If{$c$ equals to $\tau$}{
    $\theta^{'} \leftarrow \theta$ \\
    $c \leftarrow 0$}}
    \caption{Double Deep Q-learning}\label{Alg.DDQN}
\end{algorithm}\DecMargin{1em}

\section{Proposed Algorithm (R2-RLMOEA)}\label{Sec.OurAlg}
\begin{figure}[t]
\centering
\tikzstyle{decision} = [diamond, draw, fill=blue!20, text width=3.5em, text badly centered, node distance=2.2cm, inner sep=0pt]
\tikzstyle{block} = [rectangle, draw, fill=blue!20, text width=8em, text centered, rounded corners, minimum height=2em]
\tikzstyle{block1} = [rectangle, draw, fill=blue!20, text width=10em, text centered, rounded corners, minimum height=2em]
\tikzstyle{block2} = [rectangle, draw, fill=blue!20, text width=15em, text centered, rounded corners, minimum height=3em]
\tikzstyle{line} = [draw, -latex']
\tikzstyle{arrow} = [thick,->,>=stealth]
\tikzstyle{cloud} = [draw, ellipse,fill=red!20, node distance=2cm, minimum height=2em]
\scalebox{0.8}{
\begin{tikzpicture}[node distance = 1.7cm, auto]
    \node (str) [cloud] {start};
    \node (b1) [block1, below of=str, yshift=0.05cm] {\small{Initialize EAs parameters}};
    \node (b2) [block, below of=b1, yshift=0.05cm] {\small{Initialize random Pop}};
    \node (b3) [block2, below of=b2, yshift=0.05cm] {\small{Apply R2 rank operator \& update reference points}};
    \node (b4) [block, below of=b3, yshift=0.05cm] {\small{Observe states}};
    \node (b5) [block, below of=b4, yshift=0.05cm] {\small{Select an action/EA}};
    \node (b6) [block, below of=b5, yshift=0.05cm] {\small{Compute the reward}};
    \node (b7) [block, below of=b6, yshift=0.05cm] {\small{Update Q-network}};
    \node (b8) [block2, below of=b7, yshift=0.05cm] {\small{Apply R2 rank operator \& update reference points}};
    \node (b9) [decision, below of=b8, yshift=0.1cm] {\tiny{Termination}};
    \node (end) [cloud, below of=b9, yshift=0.1cm] {End};

    \path [line] (str) -- (b1);
    \path [line] (b1) -- (b2);
    \path [line] (b2) -- (b3);
    \path [line] (b3) -- (b4);
    \path [line] (b4) -- (b5);
    \path [line] (b5) -- (b6);
    \path [line] (b6) -- (b7);
    \path [line] (b7) -- (b8);
    \path [line] (b8) -- (b9);
    \path [line] (b9)node [anchor=north, xshift= 0.3cm, yshift=-0.7cm] {\small{yes}} -- (end);
    \path [line] (b9.west) node [anchor=east, xshift= -0.1cm, yshift=0.2cm] {\small{no}} -- ++ (-3.2cm,0pt) |- (b4.west);
\end{tikzpicture}}
\caption{R2-RLMOEA flow chart } \label{fig.R2RLMOEADiag}
\end{figure}
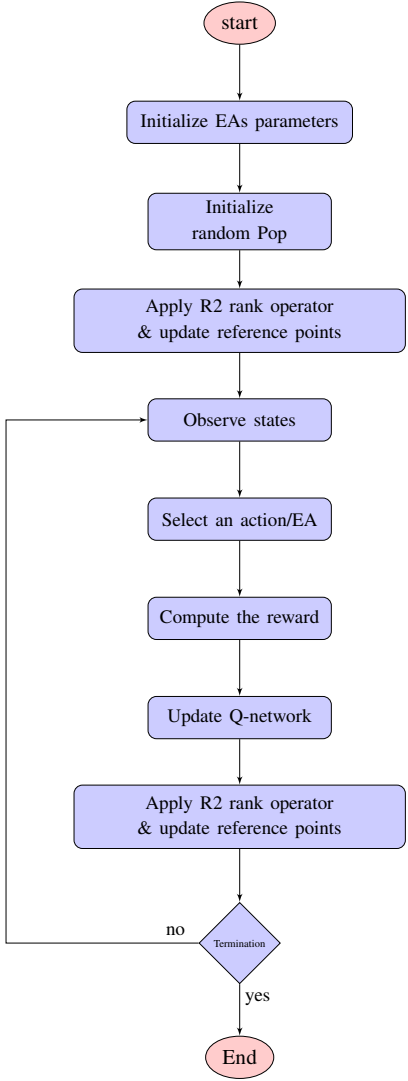

 This section presents the structure of R2-RLMOEA, including the two main EA and RL parts. The general structure of the algorithm is depicted in \myhyperref{fig.R2RLMOEADiag}{Figure}. Following the initialization of EAs and the first population, R2 ranking and updating reference points operators are applied. The RL states (the inputs of the deep neural network) are then calculated, and depending on the received inputs, the RL agent chooses an action.  Each action is associated with a specific evolutionary algorithm. Finally, the reward is predicted based on the selected EA and the current generation, and the deep network is updated. This process continues for the maximum specified number of RL iterations (maximum number of games). The final output is a trained network that can be utilized for solving a benchmark problem. 
 
The EA section includes the operators regarding the five single-objective algorithms of GA, ES, TLBO, WOA, and EO. Since we apply the single-objective form of EAs, this framework is converted to a  generic structure and other EAs might be used instead or in addition. In each generation, to convert the generated single-objective population to a multi-objective population, we apply the R2 ranking operator, explained in \myhyperref{Alg.R2}{Algorithm.}.

\IncMargin{1em}
\begin{algorithm}
\SetKwData{Left}{left}\SetKwData{This}{this}\SetKwData{Up}{up}
\SetKwFunction{Union}{Union}\SetKwFunction{FindCompress}{FindCompress}
\SetKwInOut{Input}{input}\SetKwInOut{Output}{output}
\Input{Population members ($\mathcal{P}$), and weight vector ($\mathcal{W}$)}
\Output{Sorted population members ($\mathcal{P}_{\mathrm{sorted}}$)}
\BlankLine
Compute the $L_2-\mathrm{norm}$ of $p.Obj$  ($p \in \mathcal{P}$) \\
\For{$\textbf{w} \in \mathcal{W}$}{
    \For{$p \in \mathcal{P}$}{
    $p.Sval \leftarrow ASF(p.Obj|0, \textbf{u})$   \:  \# $p.Sval$: scalarization value, \#$p.Obj$: objectives}
    $\mathcal{P}_{\mathrm{sorted}}$ = Sort($\mathcal{P}, {p.Sval}^{1st}, {L_2-\mathrm{norm}}^{2nd}$) \: \# $^{1st}$: first priority $^{2nd}$: second priority}
    \caption{Population ranking structure}\label{Alg.R2}
\end{algorithm}\DecMargin{1em}

\noindent where $p.Sval$ is the scalar value for each individual based on the ASF scalarization function, and $L_2-\mathrm{norm}$ is the Euclidean norm of all population objectives. $\mathcal{P}$ presents the current population, and the population members are ranked based on the ASF scalarization value and $L_2-\mathrm{norm}$, respectively. To improve the diversity of the solutions, we use a new reference point updating method applied in \cite{hernandez2015improved}. This method reduces the objective space and subsequently decreases the outliers.

The RL part includes two sections of online training and offline testing. For implementing DDQN, we utilize two deep networks as the main network (applied in the online and offline parts) and the target network (applied in online training). The network inputs, also called states, include twenty parameters representing the dynamic environment feature in each optimization generation. All the states are shown in \myhyperref{tab.states}{Table}

\begin{table*}[ht]
  \caption{States representation.}
  \label{tab.states}
  \center
  \setlength{\tabcolsep}{20pt} 
  \renewcommand{\arraystretch}{2} 
  \scalebox{0.9}{
  \begin{tabular}{|c|c|c|}
    \hline
    Index & State & Explanation \\
    \hline
    $s_1$ &  $\frac{f_{Q1} - f_{min}}{f_{max} - f_{min}}$ & \shortstack{\tiny{$f_{Q1}:$ lower quartile of population performance}\\ \tiny{$f_{min}$: minimum performance} \\ \tiny{$f_{max}$: maximum performance}}  \\
    \hline
    $s_2$  & $\frac{f_{Q2} - f_{min}}{f_{max} - f_{min}}$ & \tiny{$f_{Q2}$: median of population performance} \\
    \hline
    $s_3$  & $\frac{f_{Q3} - f_{min}}{f_{max} - f_{min}}$ & \tiny{$f_{Q3}$: upper quartile of population performance}\\
    \hline
    $s_4$  & $\frac{f_{Qmean} - f_{min}}{f_{max} - f_{min}}$ & \tiny{$f_{Qmean}$: average of all quartile} \\
    \hline
    $s_5$  & $\frac{\mathrm{SD}(\mathcal{P})}{\mathrm{SD}(\mathcal{P}_{minmax})}$ & \shortstack{\tiny{$\mathrm{SD}(\mathcal{P})$: population performance standard deviation}\\ \tiny{$\mathrm{SD}(\mathcal{P}_{minmax})$: maximum performance standard deviation} \\ \tiny{(half of the $\mathcal{P}_{minmax}$ includes minimum performance and the rest maximum performance)}}  \\
    \hline
    $s_6$  & $\frac{G_{max} - G_{t}}{G_{max}}$ & \shortstack{\tiny{$G_{max}$: maximum generation}\\ \tiny{$G_{t}$: current generation}}\\
    \hline
    $s_7$  & $\frac{\mathrm{ED}(x_{Q1}, x_{min})}{\mathrm{ED}(x_{max} - x_{min})}$ & \tiny{$\mathrm{ED}(.):$ Euclidean distance} \\
    \hline
    $s_8$  & $\frac{\mathrm{ED}(x_{Q2}, x_{min})}{\mathrm{ED}(x_{max} - x_{min})}$ & \shortstack{\tiny{$x_{min}$: decision variables related to $f_{min}$}\\ \tiny{$x_{max}$: decision variables related to $f_{max}$}} \\
    \hline
    $s_9$  & $\frac{\mathrm{ED}(x_{Q3}, x_{min})}{\mathrm{ED}(x_{max} - x_{min})}$ & \\
    \hline
    $s_{10}$ & $\frac{\mathrm{ED}(x_{Qmean}, x_{min})}{\mathrm{ED}(x_{max} - x_{min})}$ & \\
    \hline
    $s_{11}$ & $\frac{\mathrm{EO_{count}}}{G_{max}}$ & \tiny{$\mathrm{EO_{count}}:$ number of selected EO operator until $G_t$} \\
    \hline
    $s_{12}$ & $\frac{\mathrm{WOA_{count}}}{G_{max}}$ & \tiny{$\mathrm{WOA_{count}}:$ number of selected WOA operator until $G_t$} \\
    \hline
    $s_{13}$ & $\frac{\mathrm{TLBO_{count}}}{G_{max}}$ & \tiny{$\mathrm{TLBO_{count}}:$ number of selected TLBO operator until $G_t$} \\
    \hline
    $s_{14}$ & $\frac{\mathrm{ES_{count}}}{G_{max}}$ & \tiny{$\mathrm{ES_{count}}:$ number of selected ES operator until $G_t$} \\
    \hline
    $s_{15}$ & $\frac{\mathrm{GA_{count}}}{G_{max}}$ & \tiny{$\mathrm{GA_{count}}:$ number of selected GA operator until $G_t$} \\
    \hline
    $s_{16}$ & $\frac{\mathrm{SuEO_{count}}}{EO_{count} + \varepsilon}$ & \shortstack{\tiny{$\mathrm{SuEO_{count}}:$ number of selected EO operator that beat the previous operator} \\ \tiny{$\varepsilon: \num{1.0e-6}$}} \\
    \hline
    $s_{17}$ & $\frac{\mathrm{SuWOA_{count}}}{WOA_{count} + \varepsilon}$ & \tiny{$\mathrm{SuWOA_{count}}:$ number of selected WOA operator that beat the previous operator}\\
    \hline
    $s_{18}$ & $\frac{\mathrm{SuTlBO_{count}}}{TLBO_{count} + \varepsilon}$ & \tiny{$\mathrm{SuTLBO_{count}}:$ number of selected TLBO operator that beat the previous operator} \\
    \hline
    $s_{19}$ & $\frac{\mathrm{SuES_{count}}}{SuES_{count} + \varepsilon}$ & \tiny{$\mathrm{SuES_{count}}:$ number of selected ES operator that beat the previous operator}\\
    \hline
    $s_{20}$ & $\frac{\mathrm{SuGA_{count}}}{GA_{count} + \varepsilon}$ & \tiny{$\mathrm{SuGA_{count}}:$ number of selected GA operator that beat the previous operator} \\
    \hline
\end{tabular}}
\end{table*}

\noindent where $f_{Q1}-f_{Q3}$ represent the first, second, and third quartiles, respectively, of a current population performance. Each individual performance is calculated by adding up its $R_2$ rank and $L_2-\mathrm{norm}$, and subsequently, the overall population cost is obtained by summation of all individual performances. In each generation, the lowest quartile (for the minimization problem) demonstrates the superior performance of each algorithm. $f_{Qmean}$ also calculates the average of all quartiles. $f_{min}$ and $f_{max}$ represent the minimum and maximum performance found in each RL game up to the current EA generation ($G_t$). The $\mathcal{P}$ standard deviation of current population performance over the maximum standard deviation is calculated in $s6$. The maximum standard deviation is calculated by considering $\mathcal{P}/2$ of the population has the maximum performance (in that generation), while the rest of the population has the minimum performance. $s_7$-$s_{10}$ computes the Euclidean distance between each quartile ($s_7$-$s_9$) and the average of quartiles ($s_{10}$) with the decision variable ($x_{min}$) related to $f_{min}$. $s_{11}$-$s_{15}$ presented the number of selected specific EA in one generation over the maximum number of generations ($G_{max}$). In addition, $s_{16}$-$s_{20}$ reveals the number of a specifically selected EA, which leads to improved results (results refer to overall population cost, including the average of all population performance). The agent's criterion for evaluating the performance of each EA is the average of all quartiles. Therefore, it is considered as the reward as shown in \myhyperref{Eq.R}{Equation}.

\begin{equation}\label{Eq.R}
  Reward = \left\{ \begin{array}{lcl}
\mathcal{V} & \mbox{if} & Q(t)_{mean} > Q(t-1)_{mean} \\
  0.0 & \mbox{if} & Q(t)_{mean} \leq Q(t-1)_{mean}\end{array}\right.
\end{equation}

\noindent where $Q(t)_{mean}$ and $Q(t-1)_{mean}$ are the average of the quartiles for the current and previous generation, respectively. As the performance of the EAs in the first generations does not equal their performance in the last generations, the reward is multiplied by a value function ($\mathcal{V}$) ranging from a specific minimum to maximum value  \myhyperref{Eq.Multiple}{Equation}.

\begin{equation}\label{Eq.Multiple}
   \mathcal{V}=\left(\frac{G_{max}-G_t}{G_{max}} \right)^p \left( c_{initial} - c_{final}\right)+c_{final}
\end{equation}
\noindent where $c_{initial}$ indicates the initial reward value and $c_{final}$ specifies the final reward (for final generation). As the changes in the reward function is not linear, we also added a non-linear index ($p$) to our reward function.

\section{Experimental Settings and Evaluation metrics}\label{Sec.Setting}

To R2-RLMOEA, the CEC09 (UF1-UF10) benchmark problems with two and three objectives are considered. \myhyperref{tab.features}{Table} \cite{yang2020multi} shows the characteristics of each test problem. R2-RLMOEA is compared with the five R2-based MOEAs used in R2-RLMOEA (R2-GA [MOMBI-II], R2-ES, R2-TLBO, R2-WOA, and R2-EO) and a random MOEA in which operators are selected randomly in each generation. The comparison demonstrates how our RL-based agent performs in selecting various MOEAs during the optimization process compared to utilizing a specific algorithm and also the random selection of operators. In addition, two performance metrics, inverted generational distance (IGD) \cite{coello2005solving} and SP, are utilized to illustrate our algorithm's efficacy compared to other MOEAs.

\begin{table}[ht]
  \caption{CEC09 benchmark characteristics.}
  \label{tab.features}
  \center
  \scalebox{0.88}{
  \begin{tabular}{|c|c|c|}
    \hline
    Name & Objective/Dimension & Characteristics \\
    \hline
    UF1   & 2/30 & Concave PF, Complex PS \\
    \hline
    UF2   & 2/30 & Concave PF, Complex PS \\
    \hline
    UF3   & 2/30 & Concave PF, Complex PS \\
    \hline
    UF4   & 2/30 & Convex PF, Complex PS \\
    \hline
    UF5   & 2/30 & Discrete PF, Complex PS \\
    \hline
    UF6   & 2/30 & Discrete PF, Complex PS\\
    \hline
    UF7   & 2/30 & Complex PS \\
    \hline
    UF8   & 3/30 & Concave and Parabolic PF, Complex PS \\
    \hline
    UF9   & 3/30 & Discrete and Planar PF, Complex PS \\
    \hline
    UF10  & 3/30 & Concave and Parabolic PF \\
    \hline
\end{tabular}}
\end{table}

IGD is a popular indicator for measuring the convergence and diversity of non-dominated solutions. For a distributed reference point of $\textbf{R} = (r_1, r_2, ..., r_m)$, IGD computes the average distance of each point in set $\mathcal{P}$ to the reference point (\myhyperref{Eq.IGD}{Equation}).

\begin{align}\label{Eq.IGD}
\mathrm{IGD}(\mathcal{P}, \textbf{R}) &= \frac{\sum_{r \in \textbf{R}} \underset{x \in \mathcal{P}}{min} \sqrt{\sum_{i = 1}^{m} (f(x_i) - r_i)^2}}{|R|} 
\end{align}

\noindent where $\mathrm{ED}(.)$ is the Euclidean distance between $p \in \mathcal{P}$ with a nearest point in $\textbf{R}$ and $b$ is an arbitrary number ($b>0$). Also, we applied spacing criteria to evaluate the distribution of obtained solutions, calculating the distance (generally Euclidean distance) between each solution to its nearest neighbor in decision variable space. Based on \myhyperref{Eq.SP}{Equation}, a smaller $\mathrm{SP}$ value reveals more distributed solutions.

\begin{align}\label{Eq.SP}
\mathrm{SP} &= \frac{ \sqrt{\sum_{i = 1}^{n} (\mathcal{D}_{i} - \mathcal{D}_{m})^2}}{n \ast \mathcal{D}_{m}} 
\end{align}

\noindent where $n$ denotes the number of obtained solutions. $\mathcal{D}_i$ and $\mathcal{D}_m$ ($\sfrac{1}{n} \sum_{i = 1}^{n} \mathcal{D}_i $) are the Euclidean distance between $i^{th}$ solution and its closest neighbor and average distance, respectively. All the implementations are coded in Python. For implementing GA, ES, and some multi-objective functions, we utilized the Platypus package \cite{hadka2015platypus}. Also, the initial parameters for R2-indicator, TLBO, WOA, and EO are derived from their respective papers \cite{hernandez2015improved,rao2011teaching,mirjalili2016whale, faramarzi2020equilibrium}. The initial parameters for the General EAs and RL structure are shown in \myhyperref{tab.params}{Table}.

\begin{figure}[ht]
  \begin{center}
    \includegraphics[scale = 0.4, clip=true, trim=0cm 0cm 0cm 0cm]{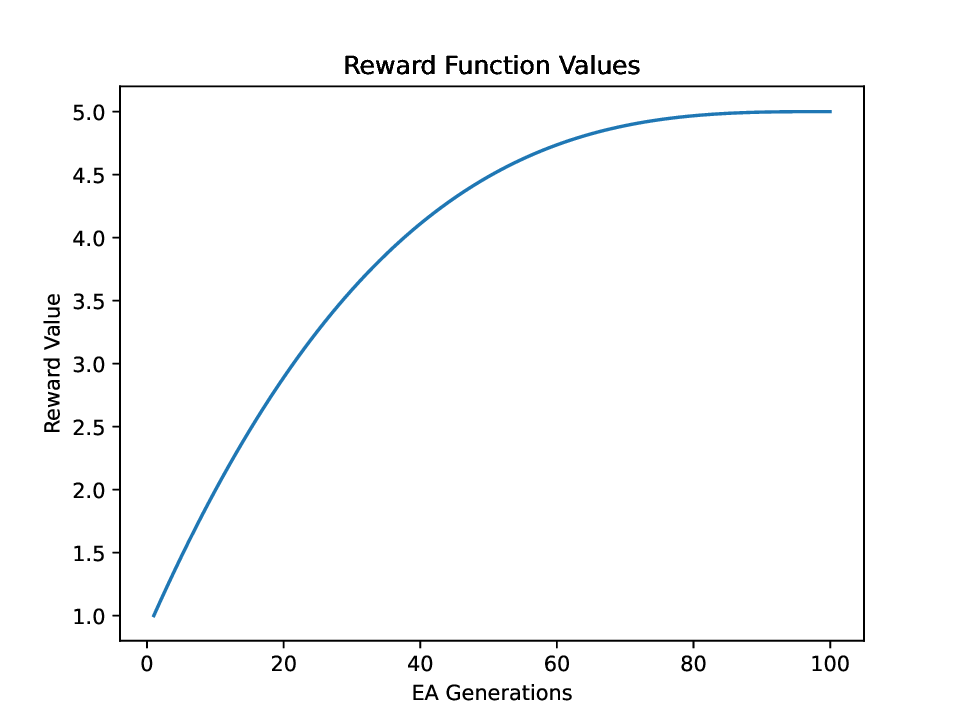}
    \caption{RL reward variation over EA generations.}
    \label{fig.RLpattern}
  \end{center}
\end{figure}

\begin{table}[ht]
  \caption{General EAs and RL parameters}
  \label{tab.params}
  \center
  \scalebox{0.85}{
  \begin{tabular}{|c c c c|}
    \hline
    \multicolumn{4}{|c|}{RL parameters} \\
    \hline
    $\mathcal{N}_{game}: \num{1.0e5}$  & $\gamma: 0.9 $ & $\mathcal{R}_{size}: $ $\num{1.0e5}$ & $n\mathcal{H}: 100$\\
    $n\mathcal{L}: 2$ & $\epsilon_{initial}: 0.9 $ & $\epsilon_{final}:\num{1.0e-3} $ & $p: 3$\\
    $\mathcal{B}_{size}: 64$ & $\mathcal{U}_{freq}:\num{1.0e3} $ &  $n_{action}: 5$ & $n_{states}: 20$\\
    \hline
    \hline
    \multicolumn{4}{|c|}{EA parameters} \\
    \hline
    $\mathcal{N}_{pop}: 100$  & $\mathcal{G}_{max}: \num{1.0e2} $ &  & \\
    \hline
\end{tabular}}
\end{table}

where $\mathcal{N}_{game}$ denotes the maximum number of RL games and $\mathcal{R}_{size}$ shows the replay memory size. In our deep model, we applied two neural networks with two hidden layers ($n\mathcal{L}$), batch size 64 ($\mathcal{B}_{size}$), and 100 nodes ($n\mathcal{H}$) in each layer. In addition, each RL game executes 100 EA generations ($\mathcal{G}_{max}$) with the population number of 100 ($\mathcal{N}_{pop}$). In RL online training, we selected actions using the epsilon greedy policy, whereas in offline mode (test mode), the agent selects actions using the greedy policy. In the epsilon-greedy policy, the epsilon varies between $\num{9.0e-1}$ to $\num{1.0e-3}$ with the power of $3$ over the game iterations. This variation pattern is also utilized in RL rewarding to make the reward function more applicable. Based on \myhyperref{fig.RLpattern}{Figure}, the reward is increased non-linearly from $1$ to $5$ over the $100$ EA generations (based on \myhyperref{Eq.R}{Equation}). 

Providing an exact computational complexity in our algorithm is challenging due to the variability in operations of each EA, as well as the dynamics of the DDQN agent. However, as a high-level approximation, we can express the overall complexity as \myhyperref{Eq.Complx}{Equation}:

\begin{align}\label{Eq.Complx}
C_{total} &= \mathcal{G}_{max} \times (C_{RL} + C_{EA}) \\
C_{total} &= \mathcal{G}_{max} \times ( O(n\mathcal{H}^2) + O(\mathcal{N}_{pop} \times \mathcal{O}_{op})) \nonumber 
\end{align}

\noindent where $C_{RL}$ is the complexity of the neural network in the RL agent and can be approximated as the $O(n\mathcal{H}^2)$. The average computational cost of the EA is $C_{EA}$, i.e., equals $O(\mathcal{N}_{pop} \times \mathcal{O}_{op})$. $\mathcal{O}_{op}$ is known as the average operations per individual in EAs, varying based on each specific EA. It is clear that by using the DDQN method, the computational complexity of our algorithm is higher than a simple MOEA. This complexity is particularly evident during the training of the network. However, since we can apply the trained network in an offline mode, the benefits of RL compensate for the higher computational training time and complexity.

After playing 80\% of the maximum number of games in each benchmark, we save the top five trained networks based on the maximum rewards they earned. These networks are then tested offline, and the ones with the best performances are reported. For each specific test function, we ran the selected network (in offline mode) 30 times for statistical analysis. 

To determine the correct statistical tests to apply, we evaluated the distribution of the average performance across all benchmarks and algorithms compared. Since a power analysis found that a sample of 28 would be required to detect a large effect at $80\%$ power and an alpha level of $0.05$, the non-parametric Friedmann test was performed to determine if there was significant difference between algorithm performance across all benchmarks, as this test is robust against outliers and does not assume a Gaussian distribution. However, to examine the performance on individual benchmarks, given the distribution of the data and the sample size, a paired mixed ANOVA is the most appropriate statistical test. For the mixed ANOVA, the between-group factor is algorithm (MOMBI-II, R2-ES, R2-TLBO, R2-WOA, R2-EO, Random Opt, and R2-RLMOEA), the within factors are benchmark (ten levels; UF1-10) and indicator (two levels; IGD and SP), and the dependent measure is the model performance (lower scores indicate better performance).

\section{Results}\label{Sec.Results}

The mean, standard deviation, and minimum of IGDs and SPs over the 30 evaluations of each algorithm are presented in \myhyperref{tab.UF}{Table}. The Friedman test results showed a significant difference in algorithm performance depending on benchmark applied and the indicator used as a measure ($\eta^2(139)=4022.61$, $p<0.001$). The highlighted cells in the tables indicate which algorithm performed the best for each benchmark.

\begin{table}[ht]
  \caption{IGD and SP metric for CEC09 test functions.}
  \label{tab.UF}
  \center
  \scalebox{0.61}{
  \begin{tabular}{|c|c|c|c|c|c|c|c|c|}
  \hline
    \multirow{7}{*}{UF1} &  & R2-RLMOEA & R2-EO & R2-WOA & R2-TLBO & R2-ES & MOMBI-II & Random $^\mathrm{a}$\\
    
   & IGD mean   & 0.10538 & 0.11867 & 0.16070 & 0.10723 & 0.77287 & 0.11075 & \cellcolor[gray]{0.7} 0.09446 \\
    
   & IGD min    & 0.08390 & 0.10397 & 0.12422 & 0.09922 & 0.56487 & 0.06794 & 0.06729 \\
    
   & IGD std    & 0.01579 & 0.01750 & 0.02035 & 0.00900 & 0.07568 & 0.03586 & 0.00954 \\
    
   & SP mean    & 0.01238 & 0.00749 & 0.01392 & 0.03036 & 0.08208 & \cellcolor[gray]{0.7} 0.00318 & 0.00814 \\
    
   & SP min     & 0.00251 & 0.00429 & 0.00538 &	0.01225 & 0.04518 &	0.00113 & 0.00210 \\
    
   & SP std     & 0.01092 & 0.00236 & 0.00945 & 0.01103 & 0.02436 & 0.00176 & 0.00430 \\
    \hline
   \multirow{7}{*}{UF2} & IGD mean   & \cellcolor[gray]{0.7} 0.04918 & 0.08924 & 0.08186 & 0.08915 & 0.64830 & 0.05011 & 0.06958\\
    
   & IGD min    & 0.03825 & 0.07783 & 0.05865 & 0.06963 & 0.55292 & 0.03805 & 0.05747 \\
    
   & IGD std    & 0.00782 & 0.00511 & 0.00758 & 0.01093 & 0.04676 & 0.01216 & 0.00740 \\
    
   & SP mean    & \cellcolor[gray]{0.7} 0.00934 & 0.02695 & 0.02568 & 0.02376 & 0.05485 & 0.01164 & 0.02024 \\
    
   & SP min     & 0.00501 & 0.01422 & 0.01031 & 0.00923 & 0.03346 & 0.00484 & 0.01261 \\
   
   & SP std     & 0.00226 & 0.00923 & 0.01147 & 0.01339 & 0.01429 & 0.00360 & 0.01141 \\
    \hline
    \multirow{7}{*}{UF3} & IGD mean   & 0.26263 & 0.37566 & 0.26473 & \cellcolor[gray]{0.7} 0.22021 & 1.13164 & 0.25866 & 0.26826 \\
    
   & IGD min    & 0.18657 & 0.30857 & 0.21376 & 0.18771 & 0.90526 & 0.21038 & 0.24050 \\
    
   & IGD std    & 0.06051 & 0.10245 & 0.02275 & 0.01992 & 0.09330 & 0.03278 & 0.01305 \\
    
   & SP mean    & \cellcolor[gray]{0.7} 0.00581 & 0.01070 & 0.01236 & 0.01059 & 0.16723 & 0.00839 & 0.00964 \\
    
   & SP min     & 0.00182 & 0.00110 & 0.00611 & 0.00602 & 0.10375 & 0.00464 & 0.00506 \\
    
   & SP std     & 0.00423 & 0.00705 & 0.00597 & 0.00454 & 0.03871 & 0.00278 & 0.00506 \\
   \hline
    \multirow{7}{*}{UF4} & IGD mean  & \cellcolor[gray]{0.7} 0.05431 & 0.09427 & 0.06804 & 0.06269 & 0.11733 & 0.05489 & 0.05619 \\
    
   &  IGD min  & 0.05055 & 0.08364 & 0.05767 & 0.05824 & 0.11082 & 0.05173 & 0.05382 \\
    
   & IGD std  & 0.00367 & 0.00531 & 0.00700 & 0.00233 & 0.00308 & 0.00331 & 0.00191 \\
    
   & SP mean  & 0.01328 & 0.01268 & 0.01882 & \cellcolor[gray]{0.7} 0.00952 & 0.01003 & 0.01606 & 0.01602 \\
    
   & SP min  & 0.00904 & 0.00982 & 0.00529 & 0.00696 & 0.00818 & 0.00893 & 0.00894 \\
    
   & SP std  & 0.00572 & 0.00201	& 0.00938 & 0.00199 & 0.00133 & 0.00609 & 0.00377 \\
   \hline
    \multirow{7}{*}{UF5} & IGD mean   & \cellcolor[gray]{0.7} 0.47041 & 0.95651 & 0.87734 & 0.96416 & 3.52099 & 0.53685 & 0.59164 \\
   
   & IGD min    & 0.24627 & 0.80740 & 0.63975 & 0.66813 & 3.07449 & 0.29841 & 0.39338 \\
    
   & IGD std    & 0.15464 & 0.09763 & 0.16326 & 0.13353 & 0.21883 & 0.14403 & 0.14759 \\
    
   & SP mean    & 0.03185 & \cellcolor[gray]{0.7} 0.02221 & 0.07299 & 0.06861 & 0.18692 & 0.02756 & 0.03620 \\
    
   & SP min     & 0.01121 & 0.00522 & 0.03673 & 0.02654 & 0.11942 &	0.01106 & 0.01902 \\
    
   & SP std     & 0.01293 & 0.02508 & 0.04388 & 0.02571 & 0.05163 &	0.01351 & 0.01079 \\
   \hline
    \multirow{7}{*}{UF6} & IGD mean  & 0.39789 & 0.54266 &	0.68166 & 0.50609 &	2.84649 & \cellcolor[gray]{0.7} 0.22817 &	0.43189 \\
    
   & IGD min  & 0.19638 & 0.45760 & 0.55416 & 0.37369 & 2.20107 & 0.12296 & 0.34400 \\
    
   & IGD std  & 0.13448 & 0.08564 & 0.05756 & 0.13473 & 0.30047 & 0.08415 & 0.06448 \\
    
   & SP mean  & 0.01232 & 0.02146 & \cellcolor[gray]{0.7} 0.01017 & 0.05508 & 0.30391 & 0.01376 & 0.02065 \\
    
   & SP min  & 0.00255 & 0.00615 & 0.00124 & 0.00785 & 0.14232 &	0.00661 & 0.00667 \\
    
   & SP std  & 0.01951 & 0.03213 & 0.01064 &	0.07354 & 0.09813 &	0.00697 & 0.02332 \\
   \hline
    \multirow{7}{*}{UF7}  & IGD mean  & \cellcolor[gray]{0.7} 0.05298 & 0.07339 & 0.10720 & 0.06002 &	0.71496 & 0.19040 &	0.05759 \\
    
   & IGD min  & 0.04209 & 0.06393 & 0.08228 & 0.05437 & 0.58447 & 0.04075 & 0.05056 \\
    
   & IGD std  & 0.00714 & 0.01273 & 0.01949 & 0.00262 & 0.06146 & 0.13883 & 0.00428 \\
    
   & SP mean  & \cellcolor[gray]{0.7} 0.00334 & 0.01354 & 0.01299 & 0.01612 & 0.08612 & 0.00432 & 0.00734 \\
    
   & SP min  & 0.00109 & 0.00420 & 0.00752 &	0.00597 & 0.05164 &	0.00059 & 0.00423 \\
    
   & SP std  & 0.00112 & 0.01096 & 0.00317 &	0.01044 & 0.02527 &	0.00595 & 0.00281 \\
   \hline
    \multirow{7}{*}{UF8}  & IGD mean  & \cellcolor[gray]{0.7} 0.28916 & 0.49866 & 0.43428 & 1.04480 &	2.31742 & 0.32214 &	0.33271 \\
    
    & IGD min  & 0.17132 & 0.39782 & 0.15952 & 0.72428 & 1.50500 & 0.17561 & 0.17740 \\
    
   & IGD std  & 0.05910 & 0.03218 & 0.09175 & 0.25355 & 0.36207 & 0.11172 & 0.11418 \\
    
   & SP mean  & 0.02619 & \cellcolor[gray]{0.7} 0.01839 & 0.04698 & 0.51333 & 0.72981 & 0.03730 & 0.05175 \\
    
   & SP min  & 0.01584 & 0.01276 & 0.01109 & 0.29071 & 0.41091 &	0.01631 & 0.01297 \\
    
   & SP std  & 0.01469 & 0.00775 & 0.04714 & 0.18741 & 0.23122 &	0.02089	 & 0.03294 \\
    \hline
    \multirow{7}{*}{UF9}  & IGD mean  & \cellcolor[gray]{0.7} 0.20521 & 0.46875 &	0.38419 & 1.25015 &	2.46484 & 0.30993 &	0.30526 \\
    
   & IGD min  & 0.09773 & 0.39060 & 0.25611 & 0.68833 & 1.63253 & 0.12381 & 0.12994 \\
    
   & IGD std  & 0.11612 & 0.04570 & 0.08392 & 0.40548 & 0.36800 & 0.12175 & 0.11141 \\
    
   & SP mean  & \cellcolor[gray]{0.7} 0.02882 & 0.18032 & 0.08126 & 0.58679 & 0.73977 & 0.04505 & 0.08789 \\
    
   & SP min  & 0.01053 & 0.05468 & 0.02350 & 0.26892 & 0.47925 &	0.00956 & 0.02428 \\
    
   & SP std  & 0.01156 & 0.11952 & 0.06534 &	0.19189 & 0.18283 &	0.03479 & 0.07486 \\
   \hline
    \multirow{7}{*}{UF10}  & IGD mean  & \cellcolor[gray]{0.7}0.42335 & 0.66104 &	0.68001 & 7.47899 &	12.90669 &	0.59200 & 1.19085 \\
    
   & IGD min  & 0.30726 & 0.39044 & 0.27059 & 4.45004 &	11.24410 & 0.32362 & 0.66454 \\
    
   & IGD std  & 0.10891 & 0.14451 & 0.33837	& 1.88916 &	0.86784	& 0.09175 &	0.36029 \\
    
   & SP mean  & \cellcolor[gray]{0.7} 0.07480 & 0.15164 & 0.14133 & 1.96243 & 2.40635 & 0.14202 & 0.20189\\
    
   & SP min  & 0.05067 & 0.07663 & 0.05280 &	1.14109 & 1.10547 &	0.05534	& 0.10491 \\
    
   & SP std  & 0.01311 & 0.04956 & 0.09954 & 0.75434 & 0.83837 &	0.06912	& 0.08863 \\
   \hline
    \multicolumn{8}{l}{$^{\mathrm{a}}$ Random Operators of EO, WOA, TLBO, ES, and GA}
\end{tabular}
}
\end{table}

Based on the results from the non-parametric analysis, the simple main effects were determined for each algorithm by comparing all algorithms with each other using a series of paired mixed ANOVA’s. For these paired mixed ANOVA’s there were enough cases to provide $80\%$ power at an alpha level of $0.05$ to detect a medium effect. The results demonstrated there was a simple main effect for the performance of the MOMBI-II algorithm compared to all others ($p < 0.001$), except the R2-RLMOEA algorithm, and for the R2-TLBO compared to the R2-ES algorithm ($p < 0.001$), and the R2-WOA algorithm compared to the R2-ES and R2-TLBO algorithms ($p < 0.001$), and the Random Opt compared to the R2-ES, R2-TLBO, R2-WOA, and R2-EO algorithms ($p < 0.03$). Overall, a simple main effect was found for the R2-RLMOEA algorithm when compared to all the other algorithms ($p < 0.001$), demonstrating that R2-RLMOEA algorithms is superior to all other when compared across all benchmarks.

Given the significantly improved performance of the R2-RLMOEA algorithm compared to all other algorithms, a series of mixed ANOVA’s were run, separately for each indicator, to evaluate differences between algorithms on each benchmark. For these ANOVA’s, the between factor was algorithm (comparing the R2-RLMOEA algorithm with each of the others, i.e., MOMBI-II, R2-ES, R2-TLBO, R2-WOA, R2-EO, and Random Opt, in a separate ANOVA), and the benchmark was the repeated measure (ten levels; UF1-10), and the dependent measure was algorithm performance on either the SP measure, or the IGD measure.

Based on SP the R2-RLMOEA algorithm significantly outperformed all other algorithms on individual benchmarks UF2, UF9, and UF10 (See \myhyperref{tab.VsSP}{Table} for a summary of pairwise comparisons). However, the R2-RLMOEA algorithm was outperformed on benchmark UF1 by the MOMBI-II algorithm ($MD = 0.008$, $SE = 0.001$, $p < 0.001$), on UF4 by the R2-ES and R2-TLBO algorithms ($MD = 0.003$, $SE = 0.009$, $p = 0.018$ and $MD = 0.005$, $SE = 0.002$, $p = 0.02$), and on UF5 by the MOMBI-II and the R2-EO algorithms ($MD = 0.01$, $SE = 0.004$, $p = 0.017$ and $MD = 0.014$, $SE = 0.003$, $p = 0.001$, respectively).

\begin{table}[ht]
  \caption{Results where the R2-RLMOEA algorithm significantly outperformed all other algorithms on benchmarks UF2, UF9, and UF10, when measured using the SP indicator.}
  \label{tab.VsSP}
  \centering
  \scalebox{0.75}{
  \begin{tabular}{|c|c|c|c|}
  \hline
    Algorithms & UF2 & UF9 & UF10 \\
    \hline
     R2-RLMOEA vs MOMBI-II & 
     \begin{tabular}{@{}c@{}}$MD = -0.004$\\ $SE = 0.001$\\ $p = 0.007$\end{tabular} & 
     \begin{tabular}{@{}c@{}}$MD = -0.124$\\ $SE = 0.035$\\ $p = 0.002$\end{tabular} & 
     \begin{tabular}{@{}c@{}}$MD = -0.062$\\ $SE = 0.011$\\ $p < 0.001$\end{tabular}\\
    \hline
     R2-RLMOEA vs R2-ES & 
     \begin{tabular}{@{}c@{}}$MD = -0.043$\\ $SE = 0.003$\\ $p < 0.001$\end{tabular} & 
     \begin{tabular}{@{}c@{}}$MD = -2.28$\\ $SE = 0.08$\\ $p < 0.001$\end{tabular} & 
     \begin{tabular}{@{}c@{}}$MD = -2.26$\\ $SE = 0.081$\\ $p < 0.001$\end{tabular} \\
    \hline
     R2-RLMOEA vs R2-TLBO & 
     \begin{tabular}{@{}c@{}}$MD = -0.009$\\ $SE = 0.003$\\ $p = 0.02$\end{tabular} & 
     \begin{tabular}{@{}c@{}}$MD = -1.12$\\ $SE = 0.099$\\ $p < 0.001$\end{tabular} & 
     \begin{tabular}{@{}c@{}}$MD = -2.032$\\ $SE = 0.23$\\ $p < 0.001$\end{tabular} \\
    \hline
     R2-RLMOEA vs R2-WOA & 
     \begin{tabular}{@{}c@{}}$MD = -0.017$\\ $SE = 0.002$\\ $p < 0.001$\end{tabular} & 
     \begin{tabular}{@{}c@{}}$MD = -0.225$\\ $SE = 0.031$\\ $p < 0.001$\end{tabular} & 
     \begin{tabular}{@{}c@{}}$MD = -0.049$\\ $SE = 0.015$\\ $p = 0.004$\end{tabular} \\
    \hline
     R2-RLMOEA vs R2-EO & 
     \begin{tabular}{@{}c@{}}$MD = -0.16$\\ $SE = 0.002$\\ $p < 0.001$\end{tabular} & 
     \begin{tabular}{@{}c@{}}$MD = -0.297$\\ $SE = 0.025$\\ $p < 0.001$\end{tabular} & 
     \begin{tabular}{@{}c@{}}$MD = -0.083$\\ $SE = 0.016$\\ $p < 0.001$\end{tabular} \\
    \hline
     R2-RLMOEA vs Random Opt & 
     \begin{tabular}{@{}c@{}}$MD = -0.007$\\ $SE = 0.001$\\ $p < 0.001$\end{tabular} & 
     \begin{tabular}{@{}c@{}}$MD = -0.16$\\ $SE = 0.032$\\ $p < 0.001$\end{tabular} & 
     \begin{tabular}{@{}c@{}}$MD = -0.1$\\ $SE = 0.009$\\ $p < 0.001$\end{tabular}  \\
    \hline
\end{tabular}
}
\end{table}

Comparing the IGD indicator as a measure of performance, the R2-RLMOEA algorithm performance was also significantly better than all other models, averaged across all ten benchmarks ($p < 0.001$). For individual benchmarks, the R2-RLMOEA algorithm performance was significantly better than all other algorithms on the benchmarks UF7, UF9, and UF10 (\myhyperref{tab.VsIGD}{Table}). However again, the R2-RLMOEA algorithm was outperformed by the MOMBI-II algorithm on benchmark UF6 ($MD = 0.166$, $SE = 0.033$, $p < 0.001$), the R2-TLBO algorithm on benchmark UF3 ($MD = 0.027$, $SE = 0.006$, $p < 0.001$), and the Random Opt algorithm on benchmark UF1 ($MD = 0.007$, $SE = 0.002$, $p = 0.007$).

\begin{table}[ht]
  \caption{Results where the R2-RLMOEA algorithm significantly outperformed all other algorithms on benchmarks UF7, UF9, and UF10, when measured using the IGD indicator.}
  \label{tab.VsIGD}
  \centering
  \scalebox{0.75}{
  \begin{tabular}{|c|c|c|c|}
  \hline
    Algorithms & UF7 & UF9 & UF10 \\
    \hline
     R2-RLMOEA vs MOMBI-II& 
     \begin{tabular}{@{}c@{}}$MD = -0.131$\\ $SE = 0.026$\\ $p < 0.001$\end{tabular} & 
     \begin{tabular}{@{}c@{}}$MD = -0.125$\\ $SE = 0.033$\\ $p = 0.001$\end{tabular} & 
     \begin{tabular}{@{}c@{}}$MD = -0.177$\\ $SE = 0.026$\\ $p < 0.001$\end{tabular} \\
    \hline
     R2-RLMOEA vs R2-ES& 
     \begin{tabular}{@{}c@{}}$MD = -0.664$\\ $SE = 0.012$\\ $p < 0.001$\end{tabular} & 
     \begin{tabular}{@{}c@{}}$MD = -2.3$\\ $SE = 0.07$\\ $p < 0.001$\end{tabular} &  
     \begin{tabular}{@{}c@{}}$MD = -12.44$\\ $SE = 0.181$\\ $p < 0.001$\end{tabular} \\
    \hline
     R2-RLMOEA vs R2-TLBO& 
     \begin{tabular}{@{}c@{}}$MD = -0.008$\\ $SE = 0.001$\\ $p < 0.001$\end{tabular} & 
     \begin{tabular}{@{}c@{}}$MD = -1.127$\\ $SE = 0.096$\\ $p < 0.001$\end{tabular} & 
     \begin{tabular}{@{}c@{}}$MD = -6.94$\\ $SE = 0.4$\\ $p < 0.001$\end{tabular} \\
    \hline
     R2-RLMOEA vs R2-WOA& 
     \begin{tabular}{@{}c@{}}$MD = -0.054$\\ $SE = 0.004$\\ $p < 0.001$\end{tabular} & 
     \begin{tabular}{@{}c@{}}$MD = -0.195$\\ $SE = 0.031$\\ $p < 0.001$\end{tabular} & 
     \begin{tabular}{@{}c@{}}$MD = -0.3$\\ $SE = 0.07$\\ $p < 0.001$\end{tabular} \\
    \hline
     R2-RLMOEA vs R2-EO& 
     \begin{tabular}{@{}c@{}}$MD = -0.017$\\ $SE = 0.001$\\ $p < 0.001$\end{tabular} & 
     \begin{tabular}{@{}c@{}}$MD = -0.278$\\ $SE = 0.016$\\ $p < 0.001$\end{tabular} & 
     \begin{tabular}{@{}c@{}}$MD = -0.272$\\ $SE = 0.045$\\ $p  < 0.001$\end{tabular} \\
    \hline
     R2-RLMOEA vs Random Opt& 
     \begin{tabular}{@{}c@{}}$MD = -0.006$\\ $SE = 0.002$\\ $p = 0.001$\end{tabular} & 
     \begin{tabular}{@{}c@{}}$MD = -0.114$\\ $SE = 0.029$\\ $p = 0.001$\end{tabular} &  
     \begin{tabular}{@{}c@{}}$MD = -0.797$\\ $SE = 0.083$\\ $p < 0.001$\end{tabular} \\
    \hline
    \end{tabular}}
\end{table}

\begin{figure}[H]
  \begin{center}
    \includegraphics[scale = 0.20, clip=true, trim=2cm 0.0cm 0.2cm 1.8cm]{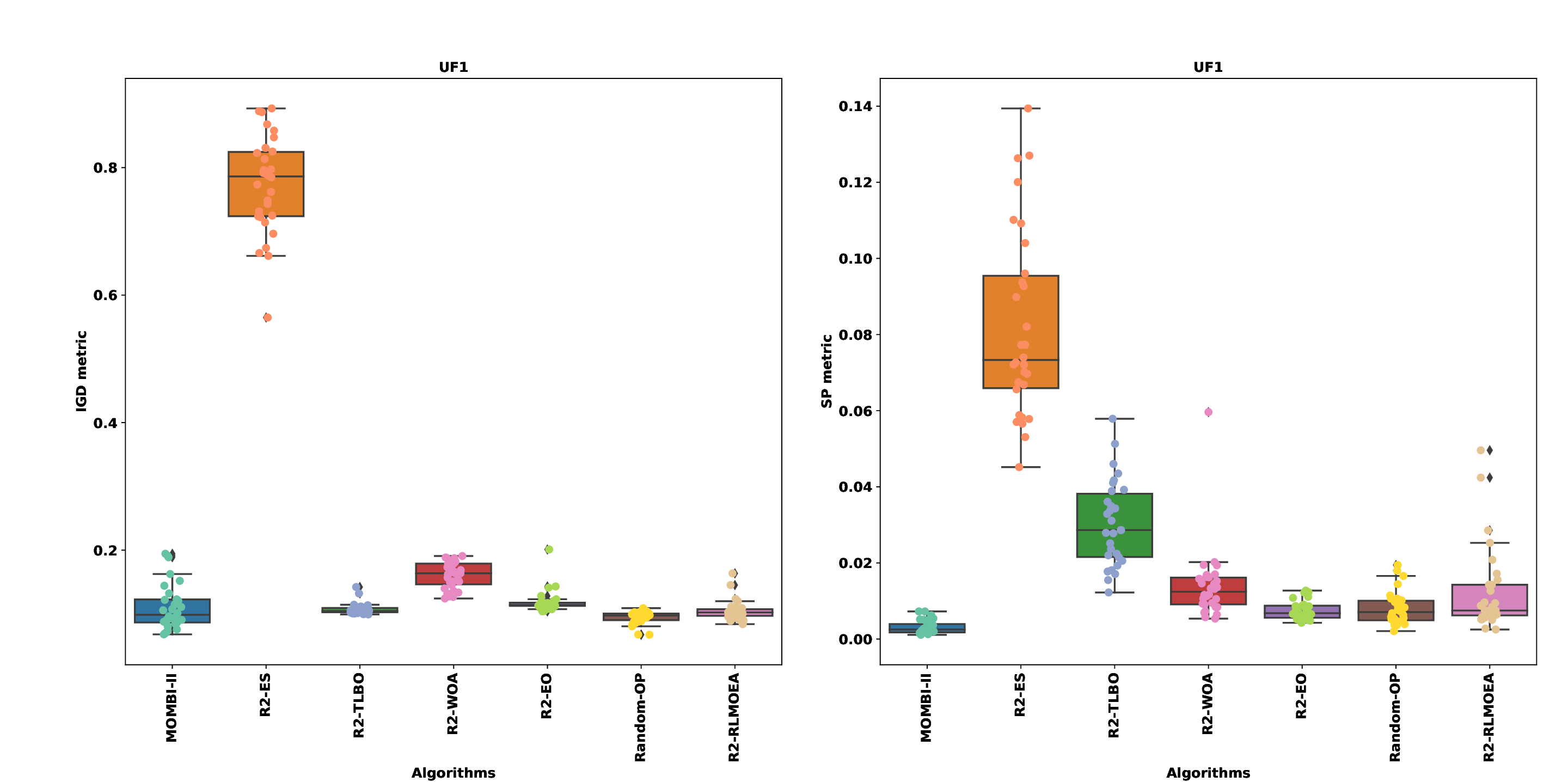}
    \caption{Box plot of the IGD and SP metrics for applied algorithms on the UF1 test function.}
    \label{fig.UF1Boxplot}
  \end{center}
\end{figure}

\begin{figure}[H]
  \begin{center}
    \includegraphics[scale = 0.20, clip=true, trim=2cm 0.0cm 0.2cm 1.8cm]{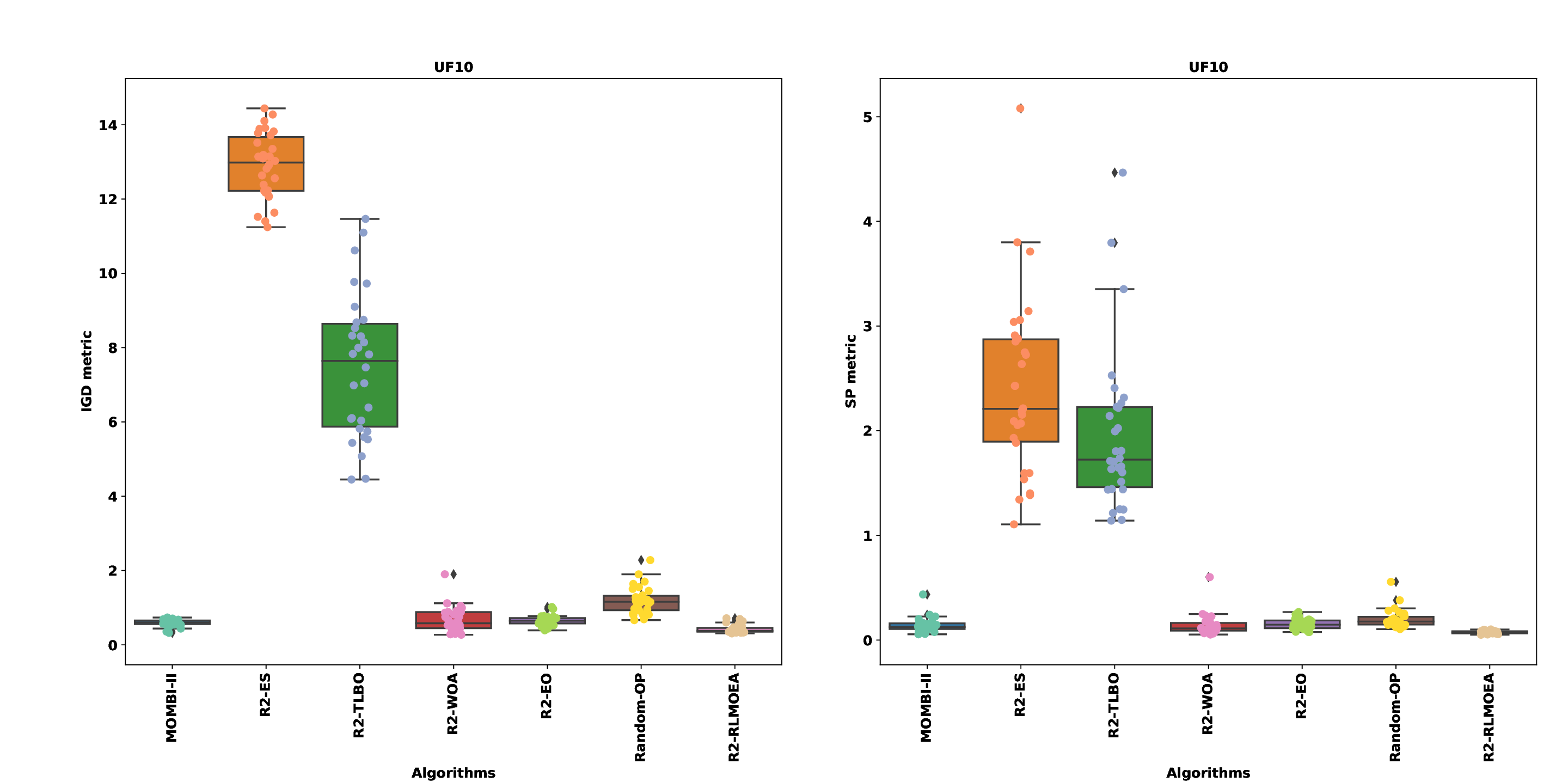}
    \caption{Box plot of the IGD and SP metrics for applied algorithms on the UF10 test function.}
    \label{fig.UF10Boxplot}
  \end{center}
\end{figure}

\begin{figure}[H]
  \begin{center}
    \includegraphics[scale = 0.20, clip=true, trim=2cm 0.0cm 0.2cm 1.8cm]{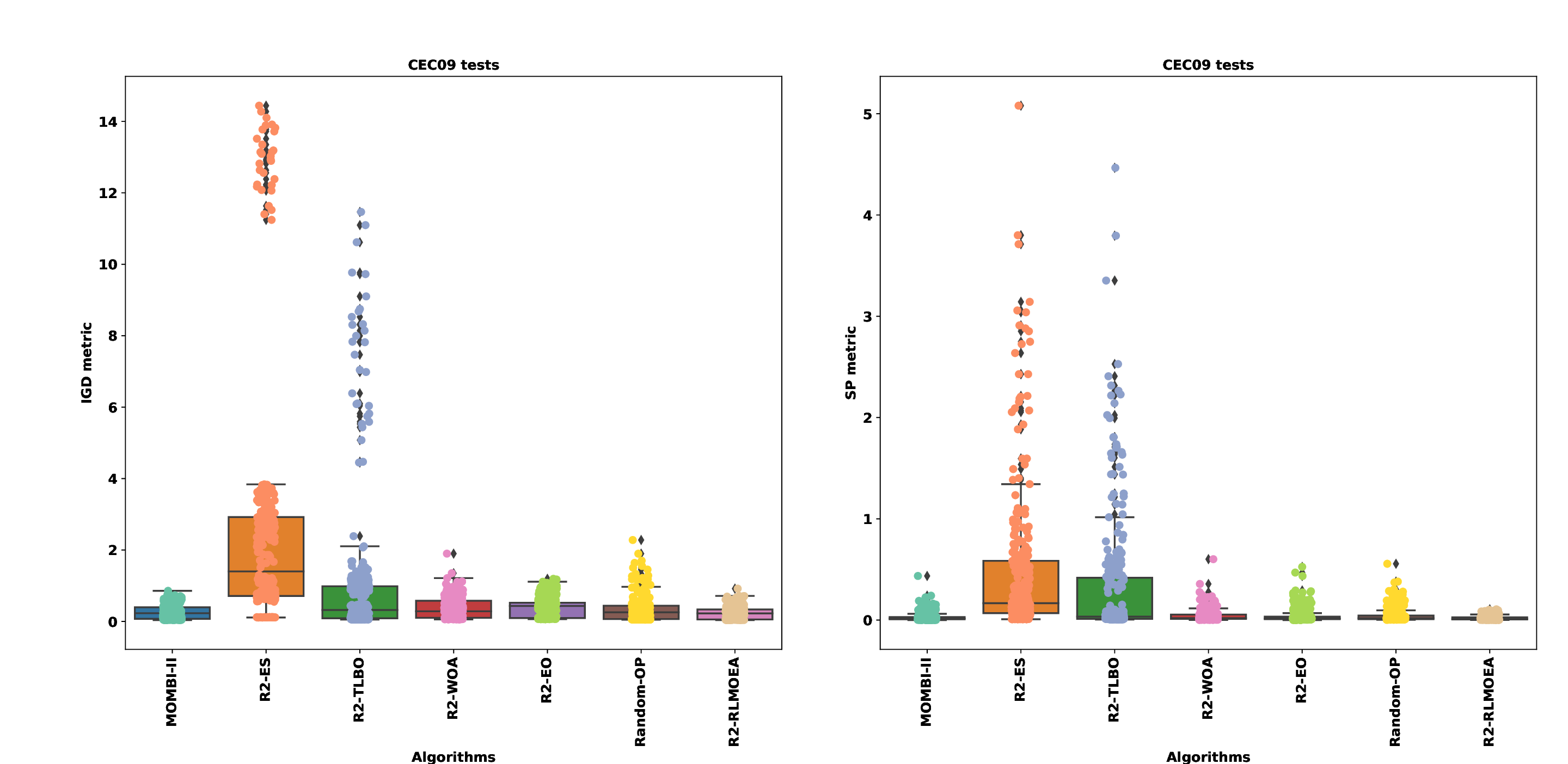}
    \caption{Combined Box plot of the IGD and SP metrics for all algorithms on the CEC09 (UF1-UF10) test functions.}
    \label{fig.Combined}
  \end{center}
\end{figure}

To provide a comprehensive visualization of the result distribution and variability among the different algorithms tested, a series of box plots are presented in Figures \myhyperref{fig.UF1Boxplot}{} to \myhyperref{fig.UF10Boxplot}{}. Each box plot represents the IGD and SP distributions (for 30 independent runs) of a specific algorithm across various benchmarks. Also, \myhyperref{fig.Combined}{Figure} displays the combined box plots for the IGD and SP indicators for all applied algorithms over the CEC09 test functions.

\begin{figure}[H]
  \begin{center}
    \includegraphics[scale = 0.20, clip=true, trim=2cm 0.0cm 0.2cm 1.8cm]{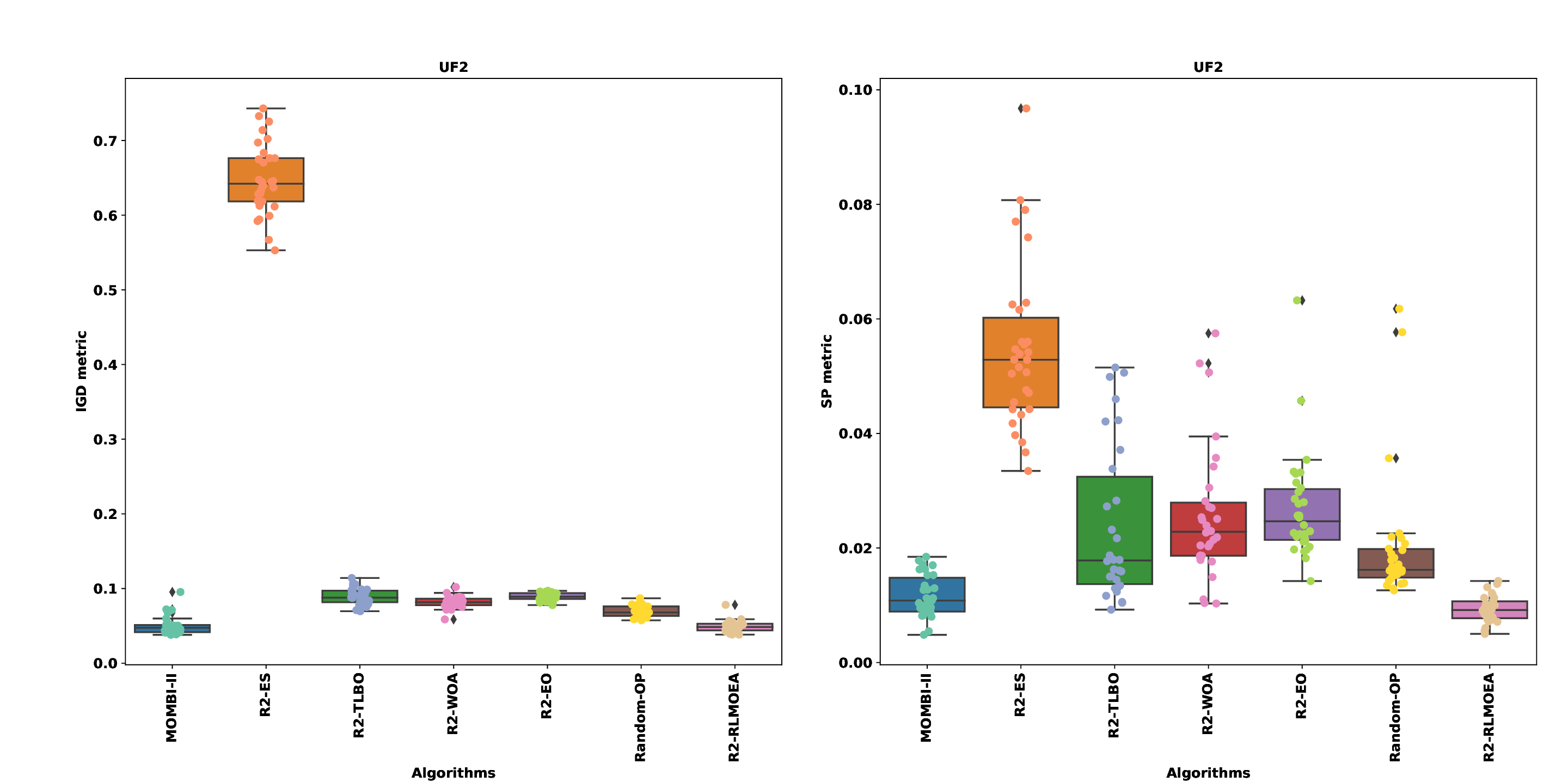}
    \caption{Box plot of the IGD and SP metrics for applied algorithms on the UF2 test function.}
    \label{fig.UF2Boxplot}
  \end{center}
\end{figure}
\begin{figure}[H]
  \begin{center}
    \includegraphics[scale = 0.20, clip=true, trim=2cm 0.0cm 0.2cm 1.8cm]{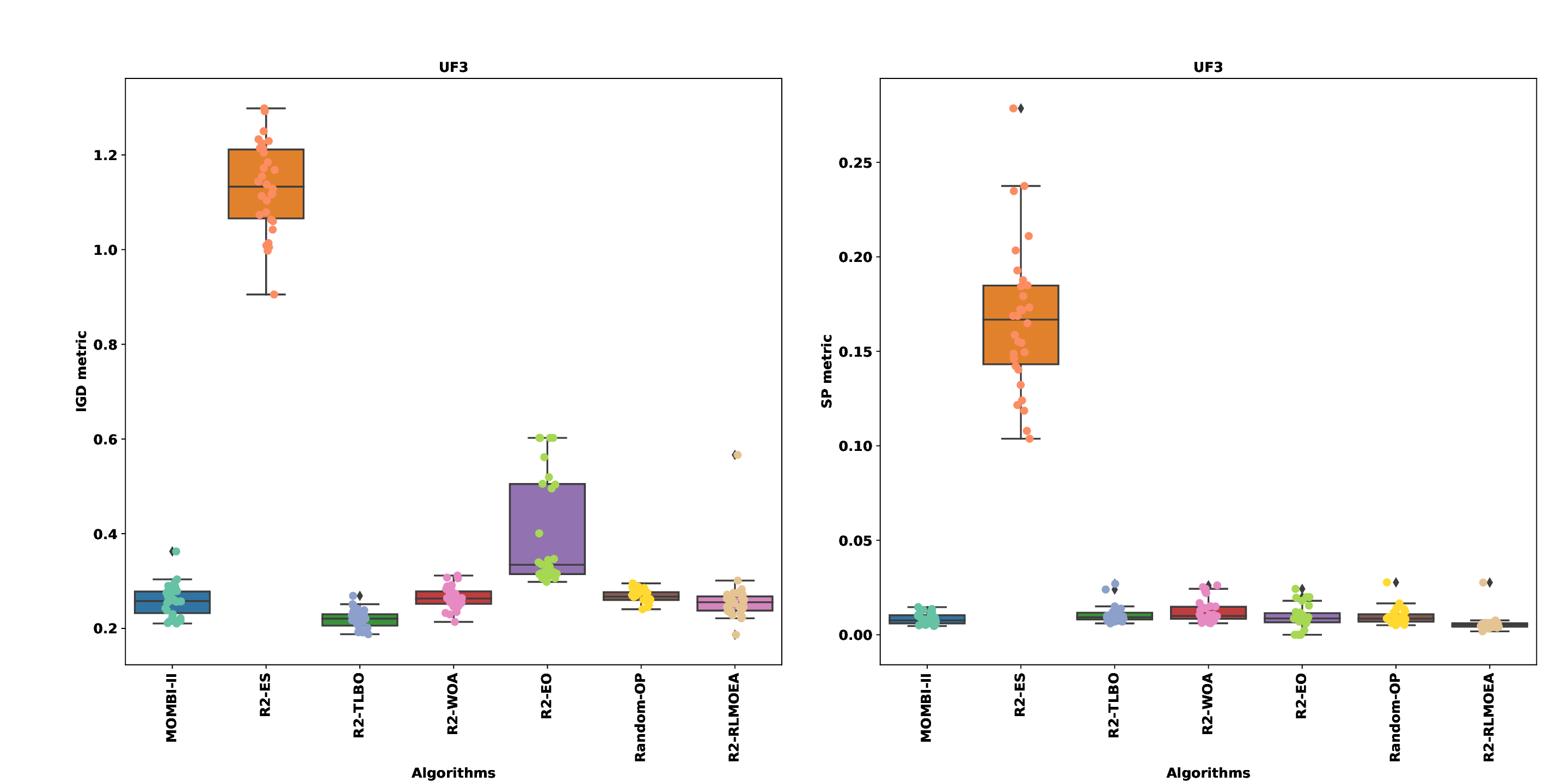}
    \caption{Box plot of the IGD and SP metrics for applied algorithms on the UF3 test function.}
    \label{fig.UF3Boxplot}
  \end{center}
\end{figure}
\begin{figure}[H]
  \begin{center}
    \includegraphics[scale = 0.20, clip=true, trim=2cm 0.0cm 0.2cm 1.8cm]{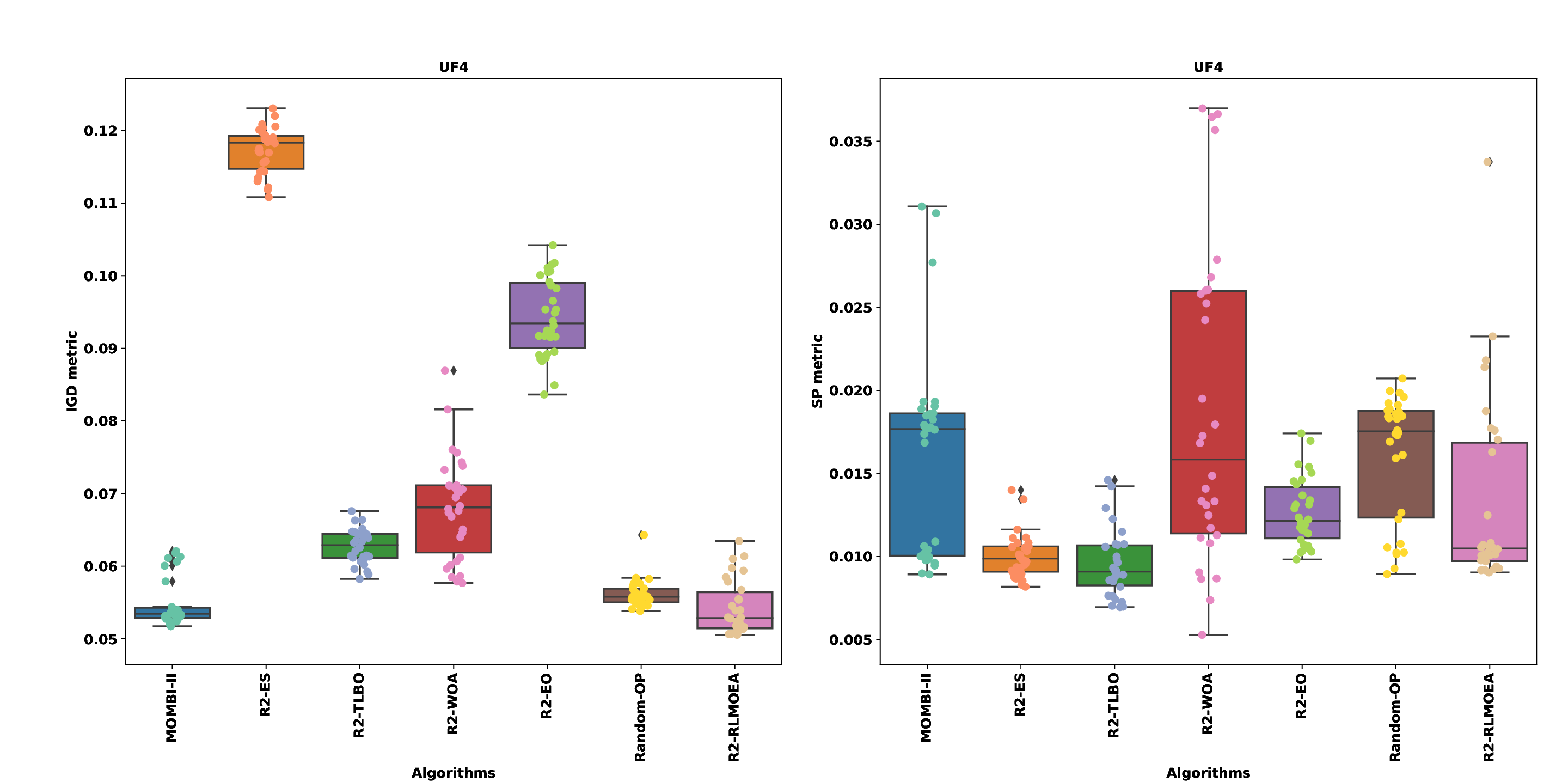}
    \caption{Box plot of the IGD and SP metrics for applied algorithms on the UF4 test function.}
    \label{fig.UF4Boxplot}
  \end{center}
\end{figure}

\begin{figure}[H]
  \begin{center}
    \includegraphics[scale = 0.20, clip=true, trim=2cm 0.0cm 0.2cm 1.8cm]{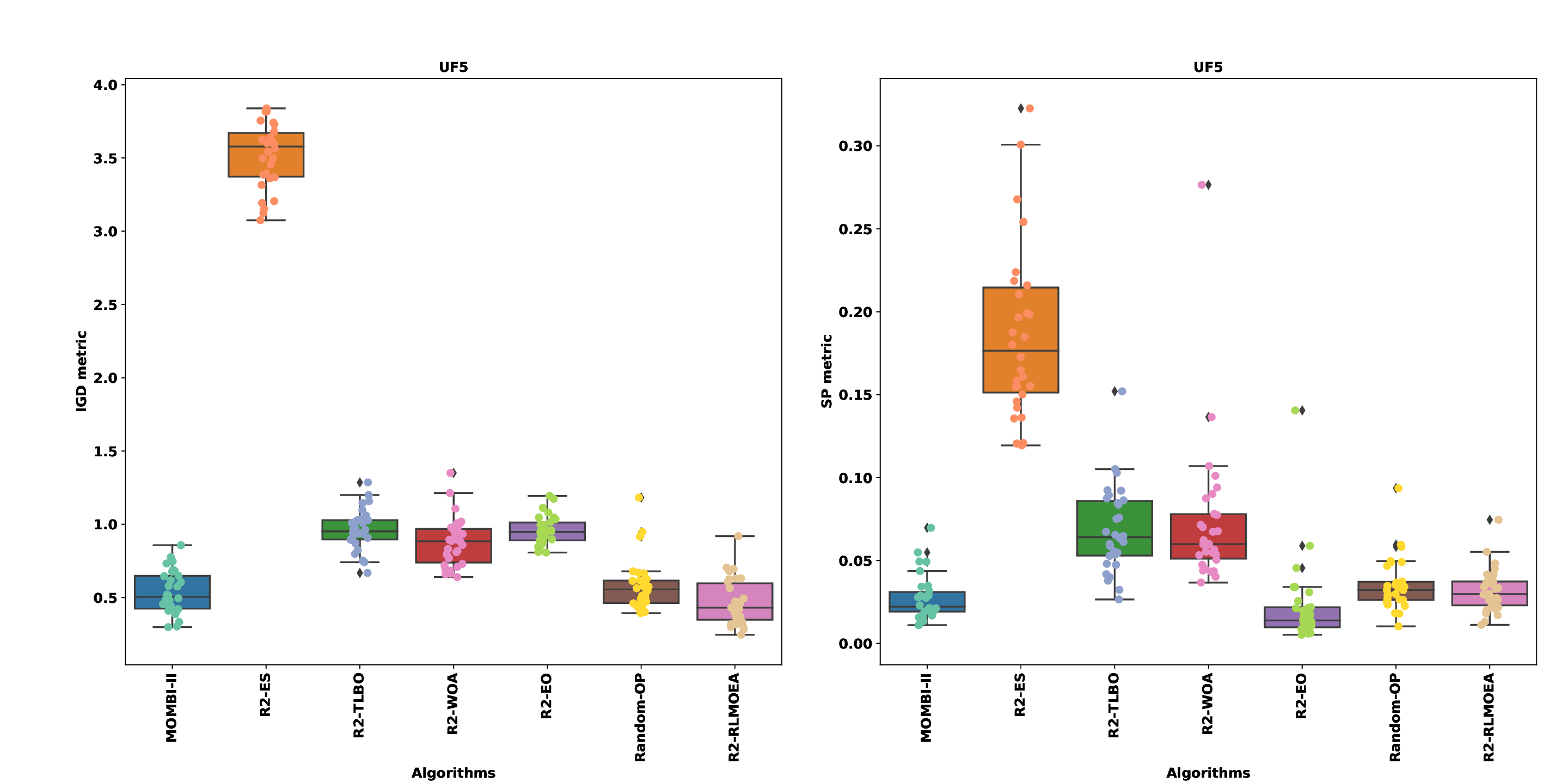}
    \caption{Box plot of the IGD and SP metrics for applied algorithms on the UF5 test function.}
    \label{fig.UF5Boxplot}
  \end{center}
\end{figure}

\begin{figure}[H]
  \begin{center}
    \includegraphics[scale = 0.20, clip=true, trim=2cm 0.0cm 0.2cm 1.8cm]{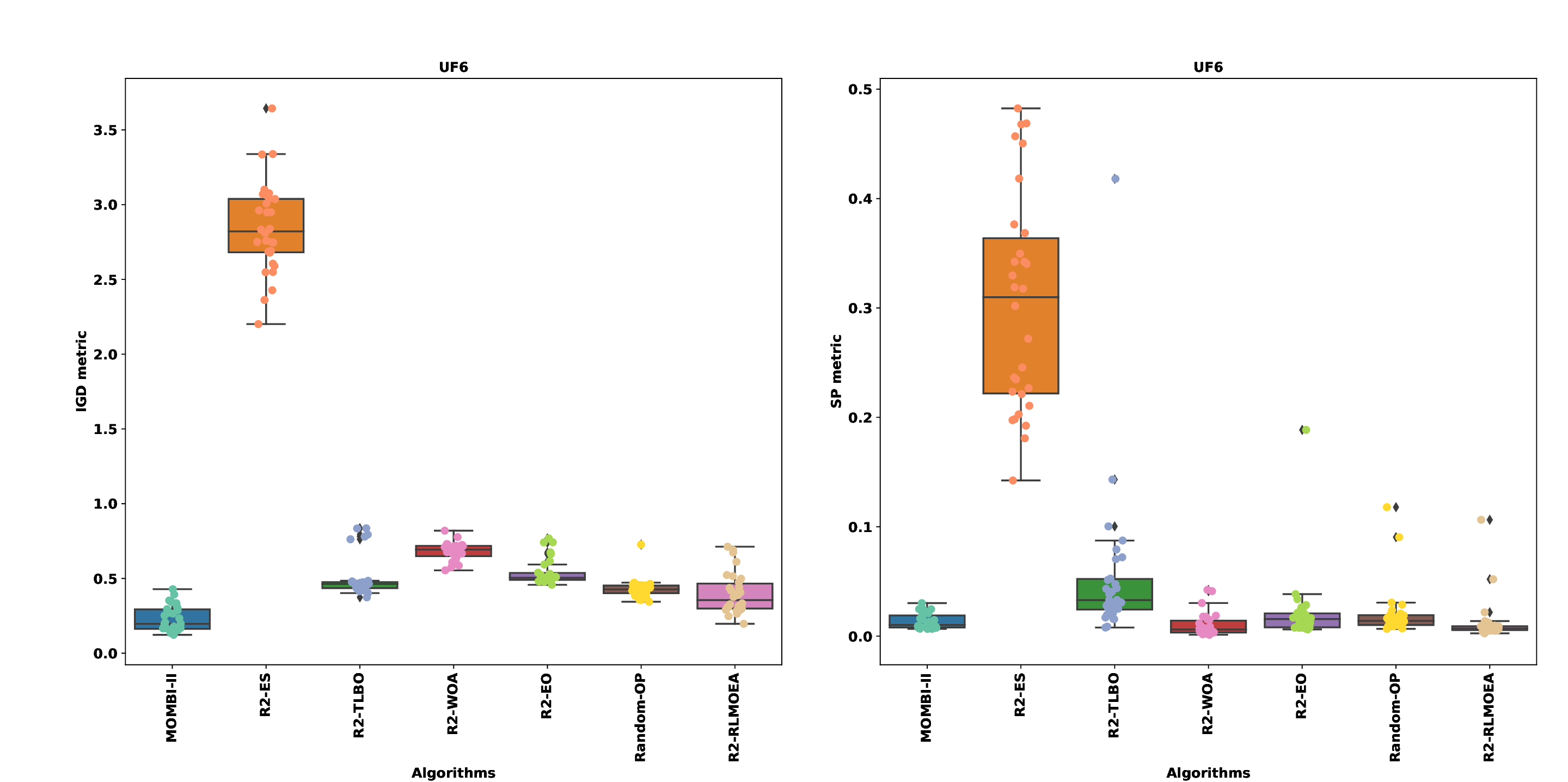}
    \caption{Box plot of the IGD and SP metrics for applied algorithms on the UF6 test function.}
    \label{fig.UF6Boxplot}
  \end{center}
\end{figure}

\begin{figure}[H]
  \begin{center}
    \includegraphics[scale = 0.20, clip=true, trim=2cm 0.0cm 0.2cm 1.8cm]{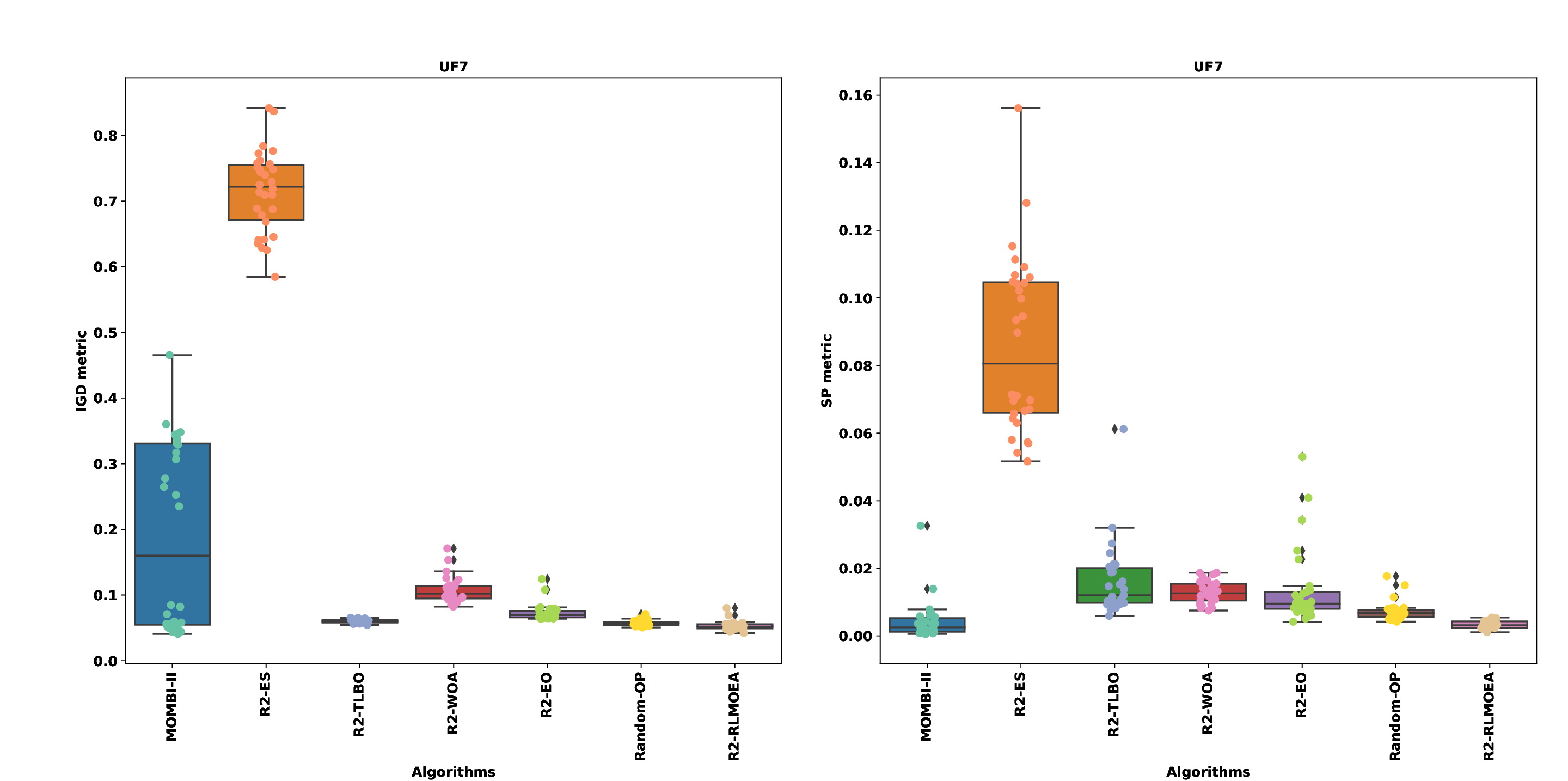}
    \caption{Box plot of the IGD and SP metrics for applied algorithms on the UF7 test function.}
    \label{fig.UF7Boxplot}
  \end{center}
\end{figure}

\begin{figure}[H]
  \begin{center}
    \includegraphics[scale = 0.20, clip=true, trim=2cm 0.0cm 0.2cm 1.8cm]{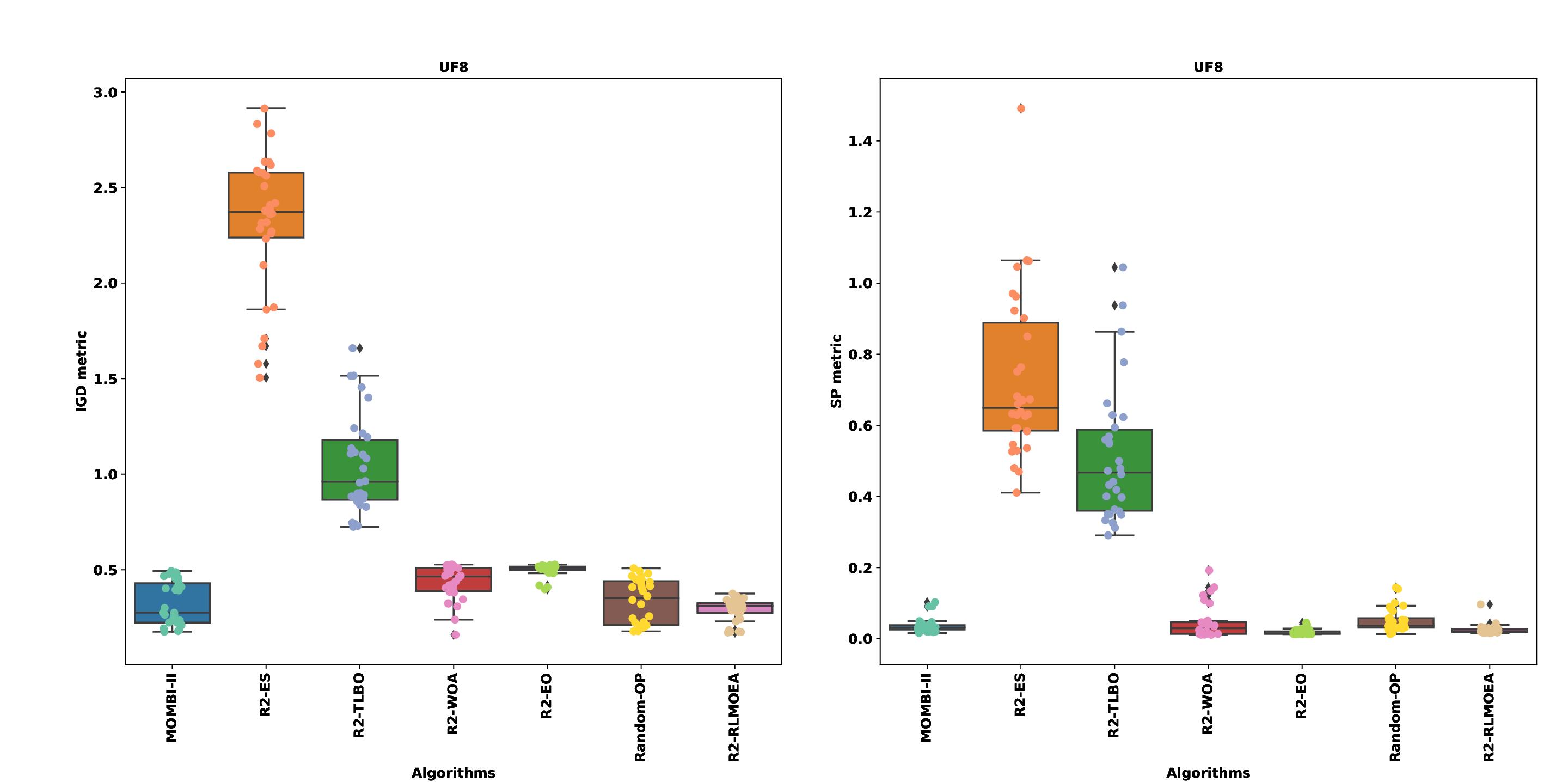}
    \caption{Box plot of the IGD and SP metrics for applied algorithms on the UF8 test function.}
    \label{fig.UF8Boxplot}
  \end{center}
\end{figure}

\begin{figure}[H]
  \begin{center}
    \includegraphics[scale = 0.20, clip=true, trim=2cm 0.0cm 0.2cm 1.8cm]{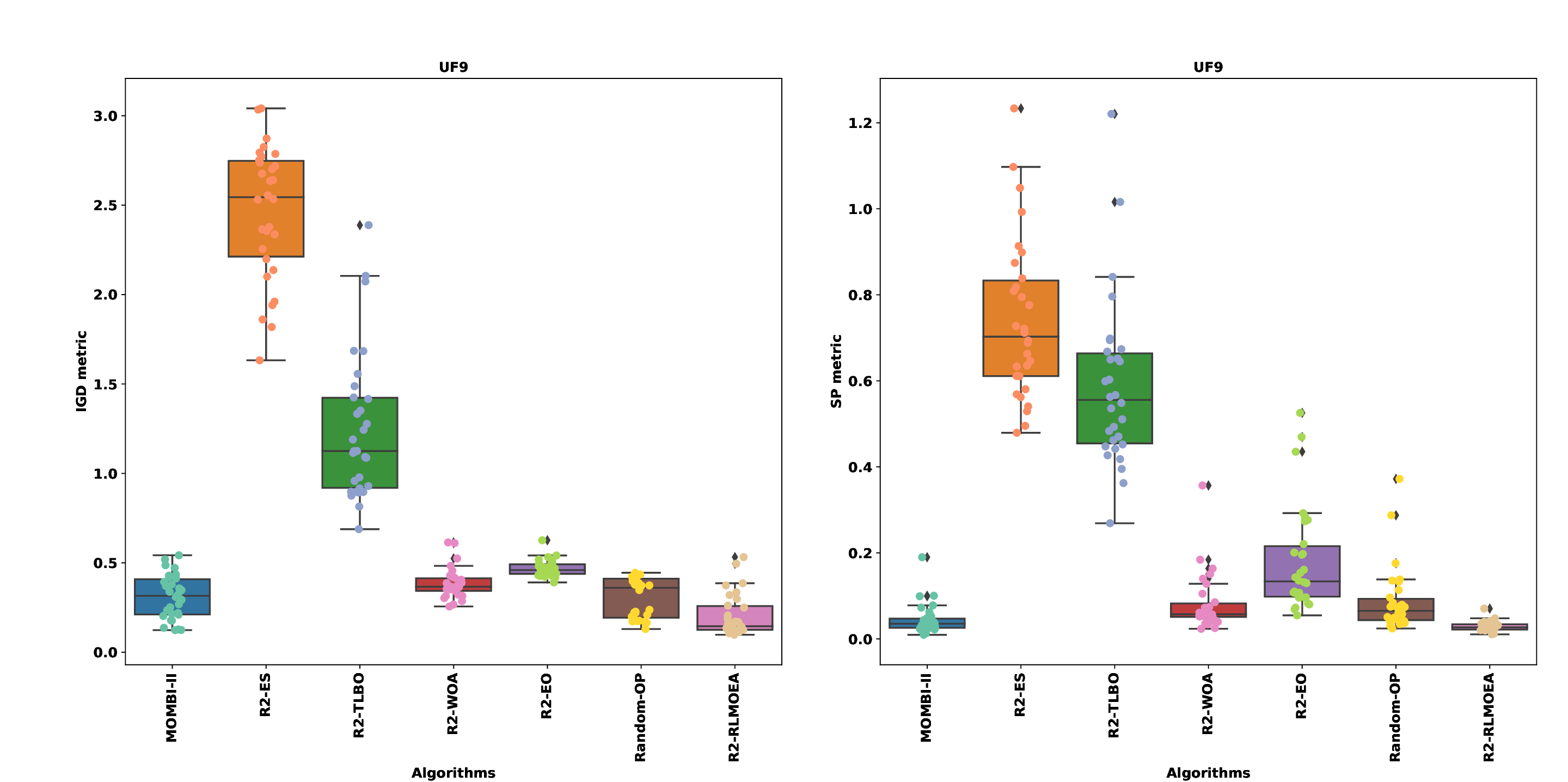}
    \caption{Box plot of the IGD and SP metrics for applied algorithms on the UF9 test function.}
    \label{fig.UF9Boxplot}
  \end{center}
\end{figure}

\section{EA Choice Assessment} \label{Sec.EAChoice}
\begin{figure}[H]
  \begin{center}
    \includegraphics[scale = 0.19, clip=true, trim=0.1cm 0.3cm 0.1cm 0.3cm]{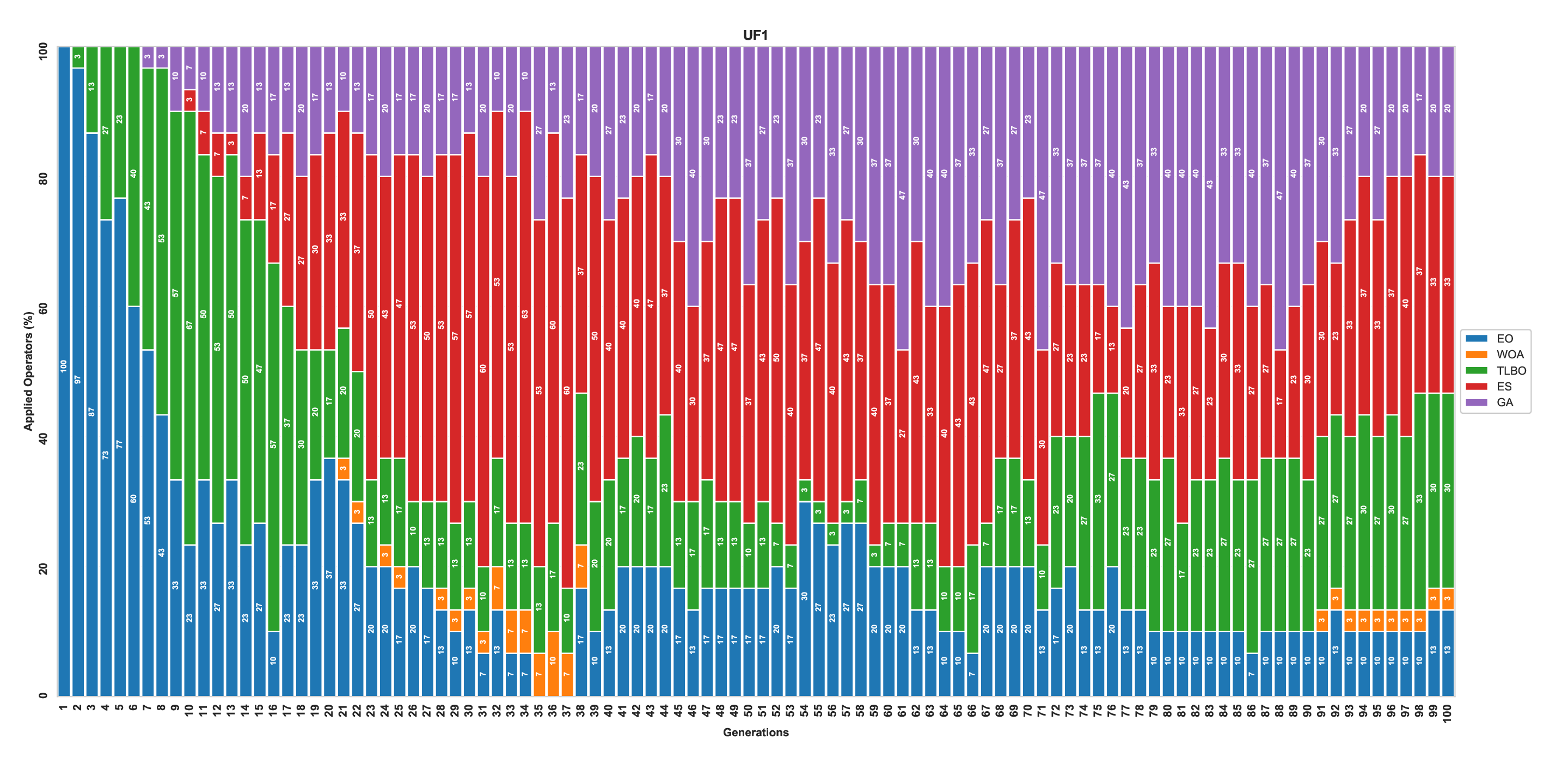}
    \caption{The percentage of each EA selected by the RL agent in each generation for the UF1 test function.}
    \label{fig.UF1Algplot}
  \end{center}
\end{figure}

\begin{figure}[ht]
  \begin{center}
    \includegraphics[scale = 0.19, clip=true, trim=0.1cm 0.3cm 0.1cm 0.3cm]{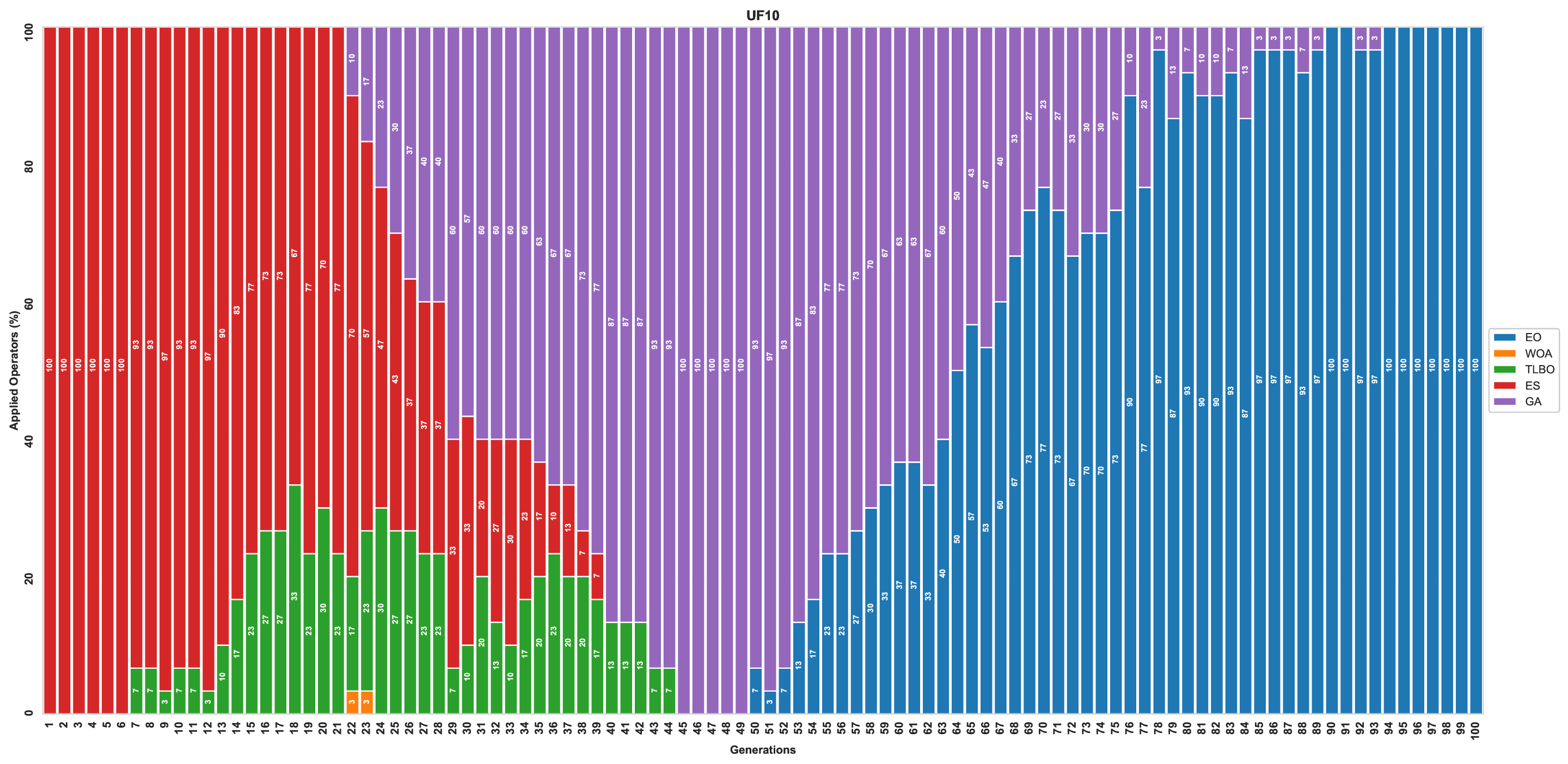}
    \caption{The percentage of each EA selected by the RL agent in each generation for the UF10 test function.}
    \label{fig.UF10Algplot}
  \end{center}
\end{figure}

Selecting the appropriate evolutionary algorithm (EA) for a particular problem is crucial and requires professional expertise. In our framework, we take this a step further by selecting a specific EA for each generation during the optimization process. By analyzing over 30 independent runs, we reveal the percentage of selection of each algorithm in specific generations. The results for operator selection for UF1 to UF10 are presented in Figures \myhyperref{fig.UF1Algplot}{} to \myhyperref{fig.UF10Algplot}{}, respectively. Based on the results, it can be concluded that the agent prefers using the ES method during the initial generations due to its exploration capabilities. However, as the optimization process progresses and requires more exploitation of the search space, the agent switches to GA and TLBO in the middle generations, and EO and WOA in the last generations, in most cases. This trend was particularly noticeable when dealing with three-objective test functions. Optimisation may be suboptimal due to the agent's lack of confidence or poor selection of operators with respect to the exploration and exploitation criteria. In addition, the percentage contributions of various EA operators to a variety of test problems are displayed in \myhyperref{tab.operatorcontribution}{Table}, demonstrating how the selection process adapts to the specific characteristics of each problem. Despite the diverse range of operators utilized throughout these benchmarks, it is evident that the GA and TLBO operators predominate as the operators utilized in the CEC09 benchmark.

\begin{figure}[H]
  \begin{center}
    \includegraphics[scale = 0.19, clip=true, trim=0.1cm 0.3cm 0.1cm 0.3cm]{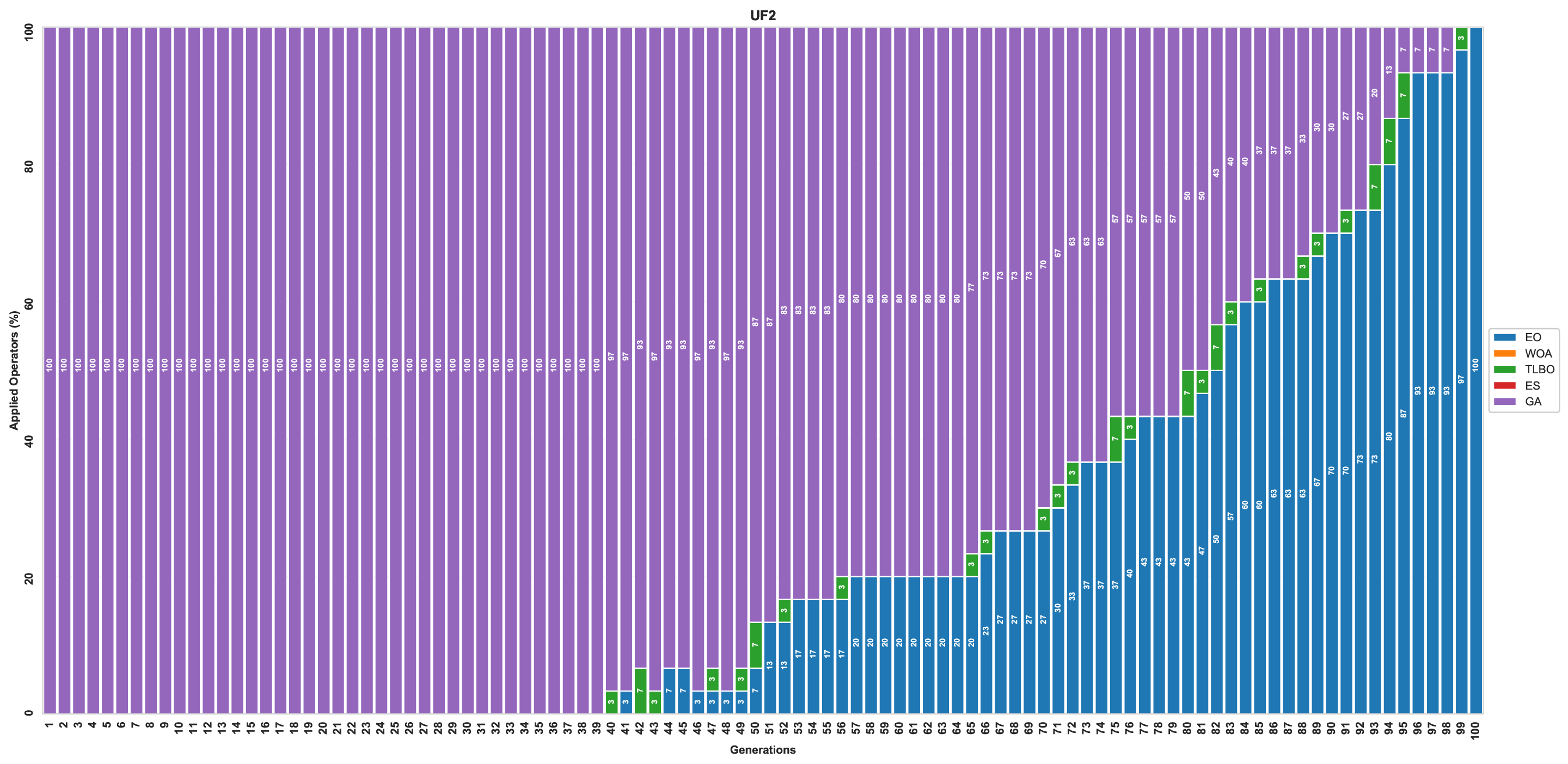}
    \caption{The percentage of each EA selected by the RL agent in each generation for the UF2 test function.}
    \label{fig.UF2Algplot}
  \end{center}
\end{figure}
\begin{figure}[H]
  \begin{center}
    \includegraphics[scale = 0.19, clip=true, trim=0.1cm 0.3cm 0.1cm 0.3cm]{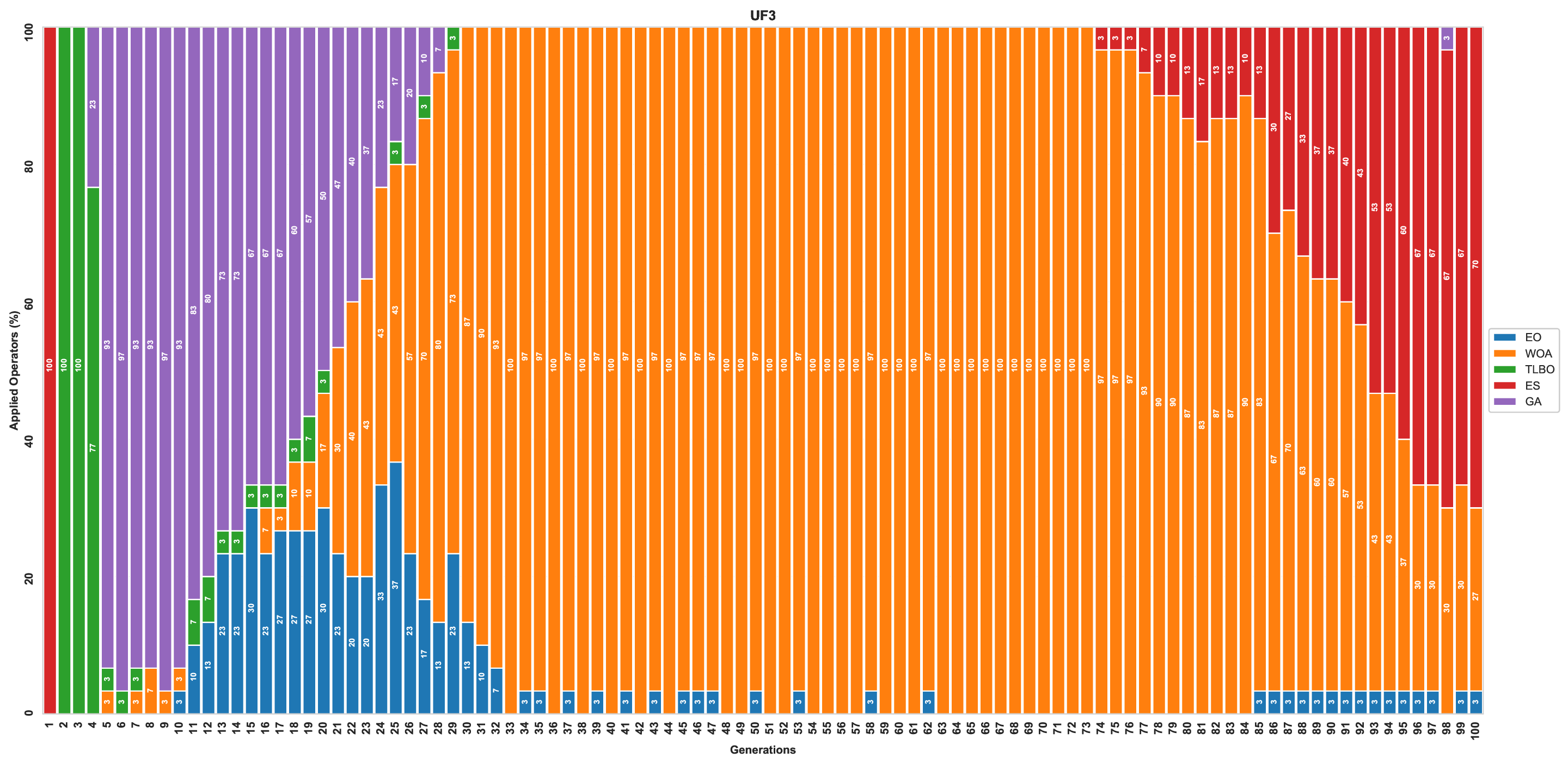}
    \caption{The percentage of each EA selected by the RL agent in each generation for the UF3 test function.}
    \label{fig.UF3Algplot}
  \end{center}
\end{figure}
\begin{figure}[H]
  \begin{center}
    \includegraphics[scale = 0.19, clip=true, trim=0.1cm 0.3cm 0.1cm 0.3cm]{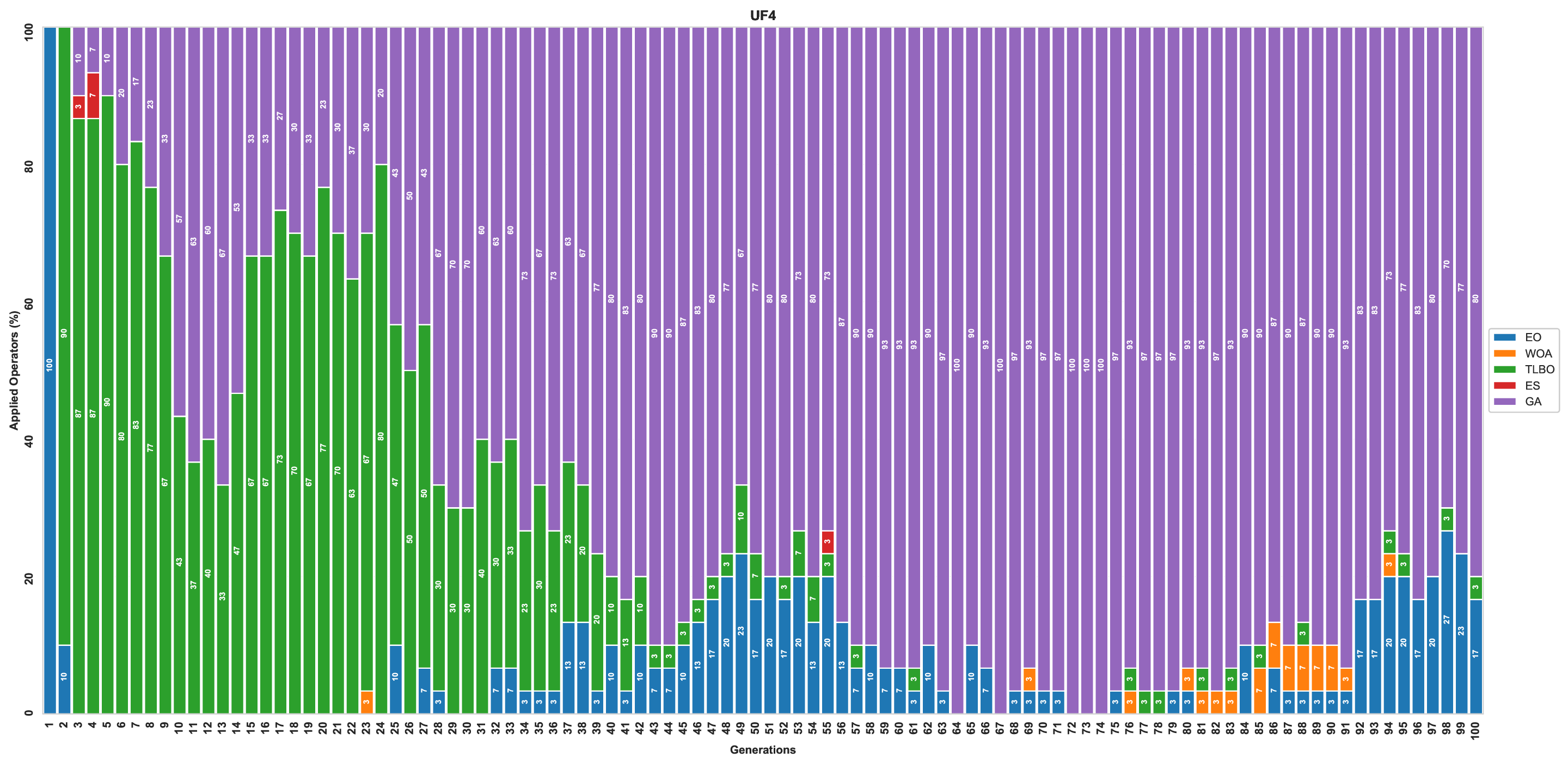}
    \caption{The percentage of each EA selected by the RL agent in each generation for the UF4 test function.}
    \label{fig.UF4Algplot}
  \end{center}
\end{figure}

\begin{figure}[H]
  \begin{center}
    \includegraphics[scale = 0.19, clip=true, trim=0.1cm 0.3cm 0.1cm 0.3cm]{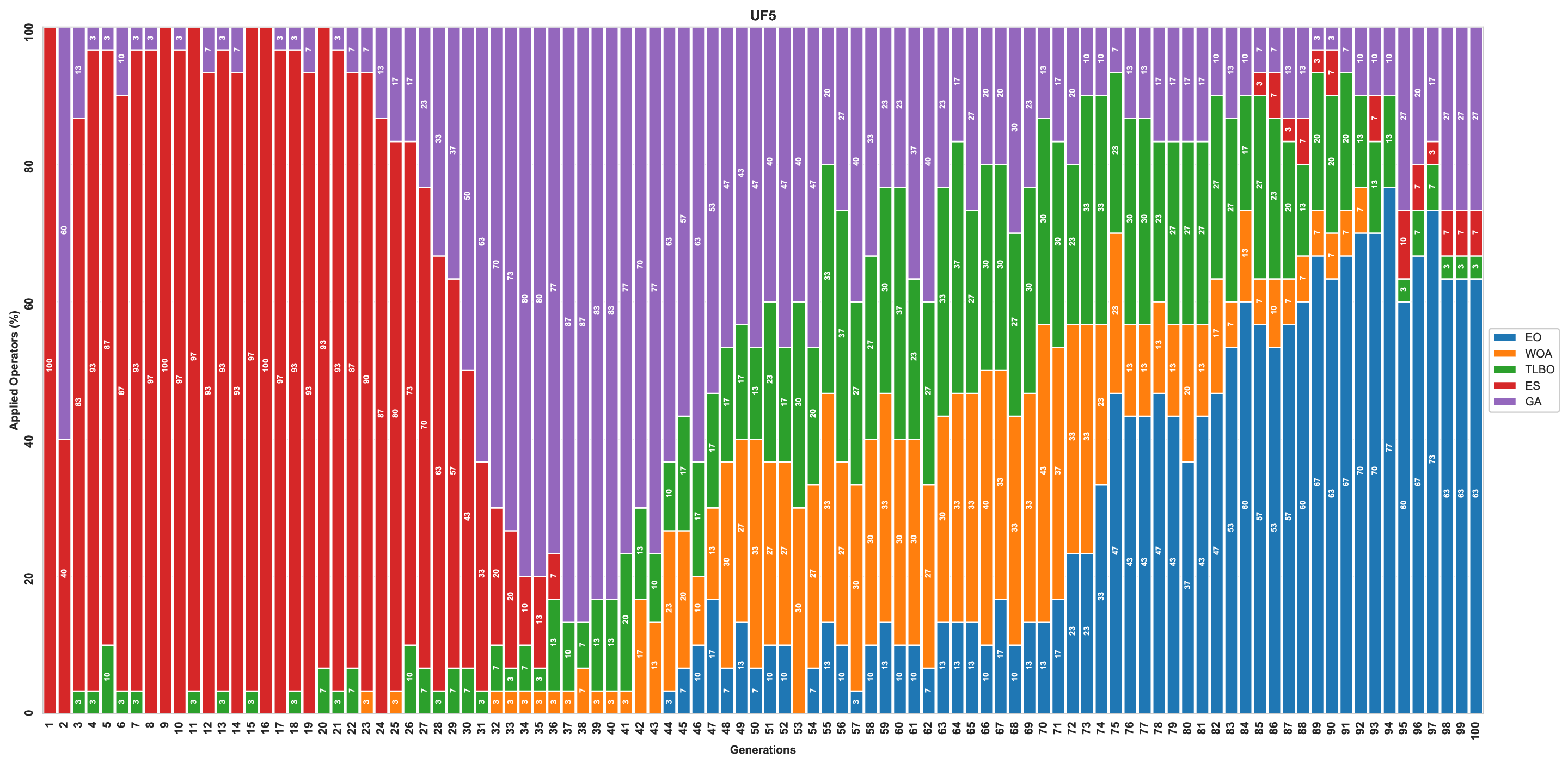}
    \caption{The percentage of each EA selected by the RL agent in each generation for the UF5 test function.}
    \label{fig.UF5Algplot}
  \end{center}
\end{figure}

\begin{figure}[H]
  \begin{center}
    \includegraphics[scale = 0.19, clip=true, trim=0.1cm 0.3cm 0.1cm 0.3cm]{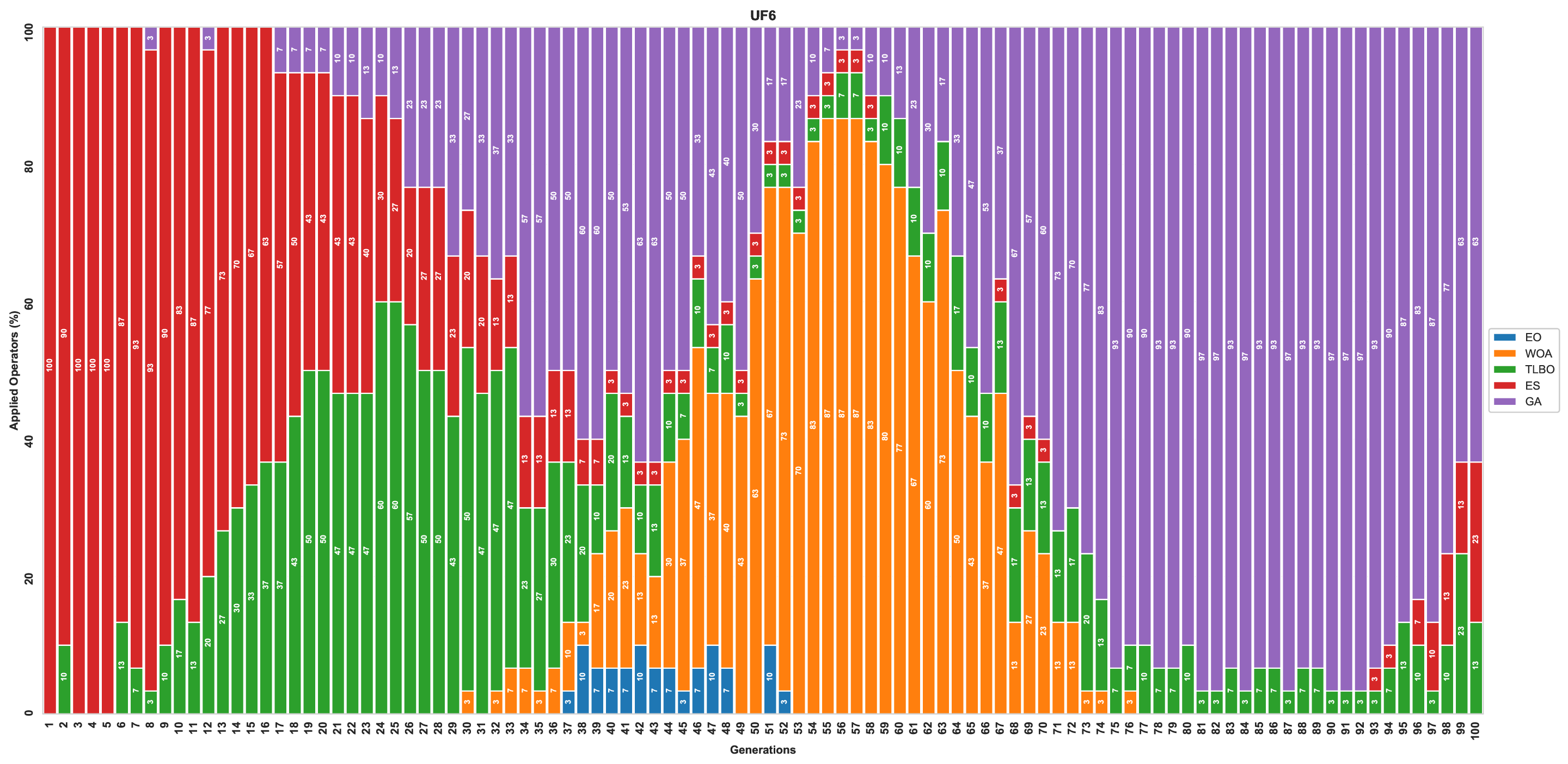}
    \caption{The percentage of each EA selected by the RL agent in each generation for the UF6 test function.}
    \label{fig.UF6Algplot}
  \end{center}
\end{figure}

\begin{figure}[H]
  \begin{center}
    \includegraphics[scale = 0.19, clip=true, trim=0.1cm 0.3cm 0.1cm 0.3cm]{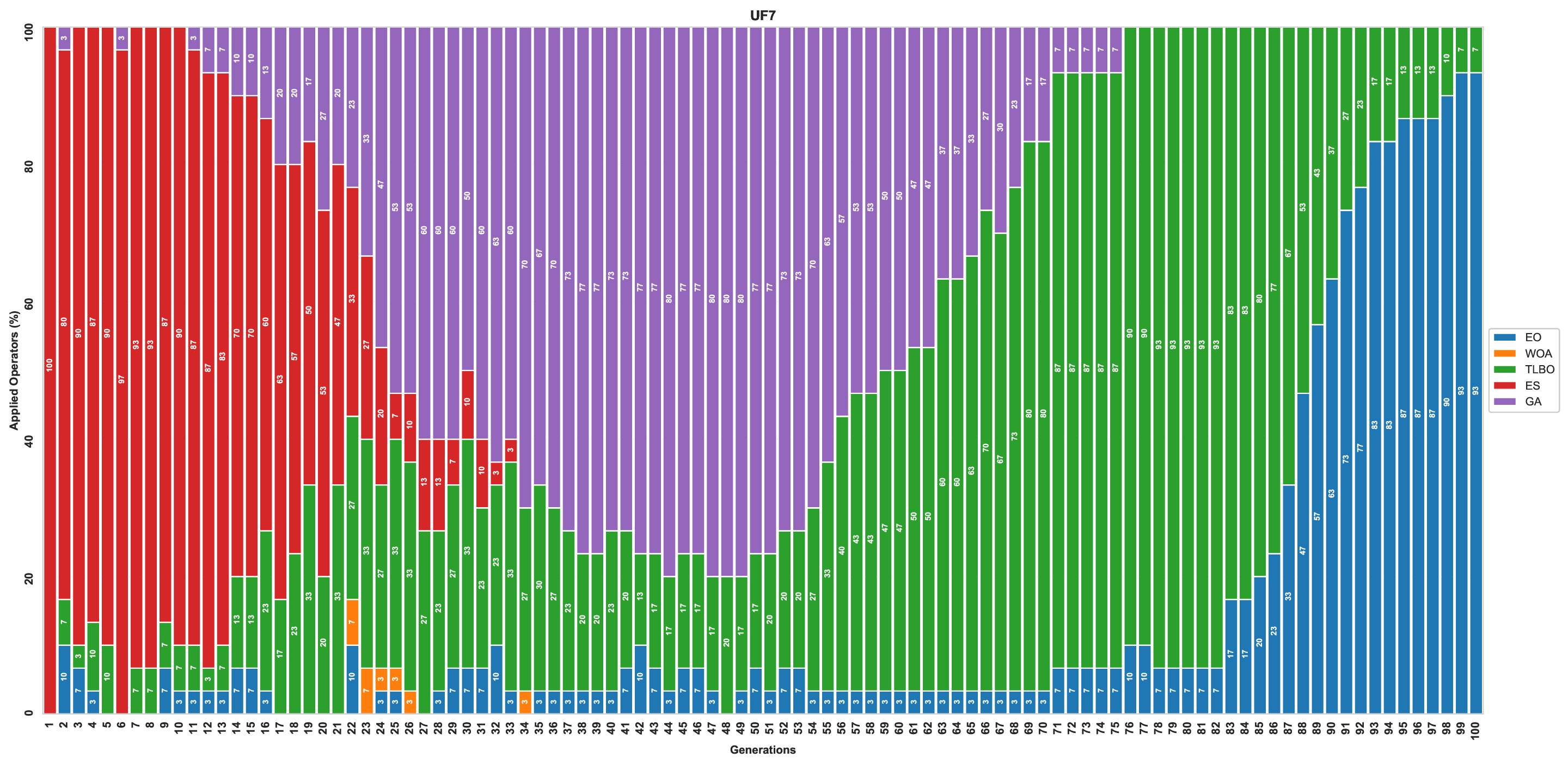}
    \caption{The percentage of each EA selected by the RL agent in each generation for the UF7 test function.}
    \label{fig.UF7Algplot}
  \end{center}
\end{figure}

\begin{figure}[H]
  \begin{center}
    \includegraphics[scale = 0.19, clip=true, trim=0.1cm 0.3cm 0.1cm 0.3cm]{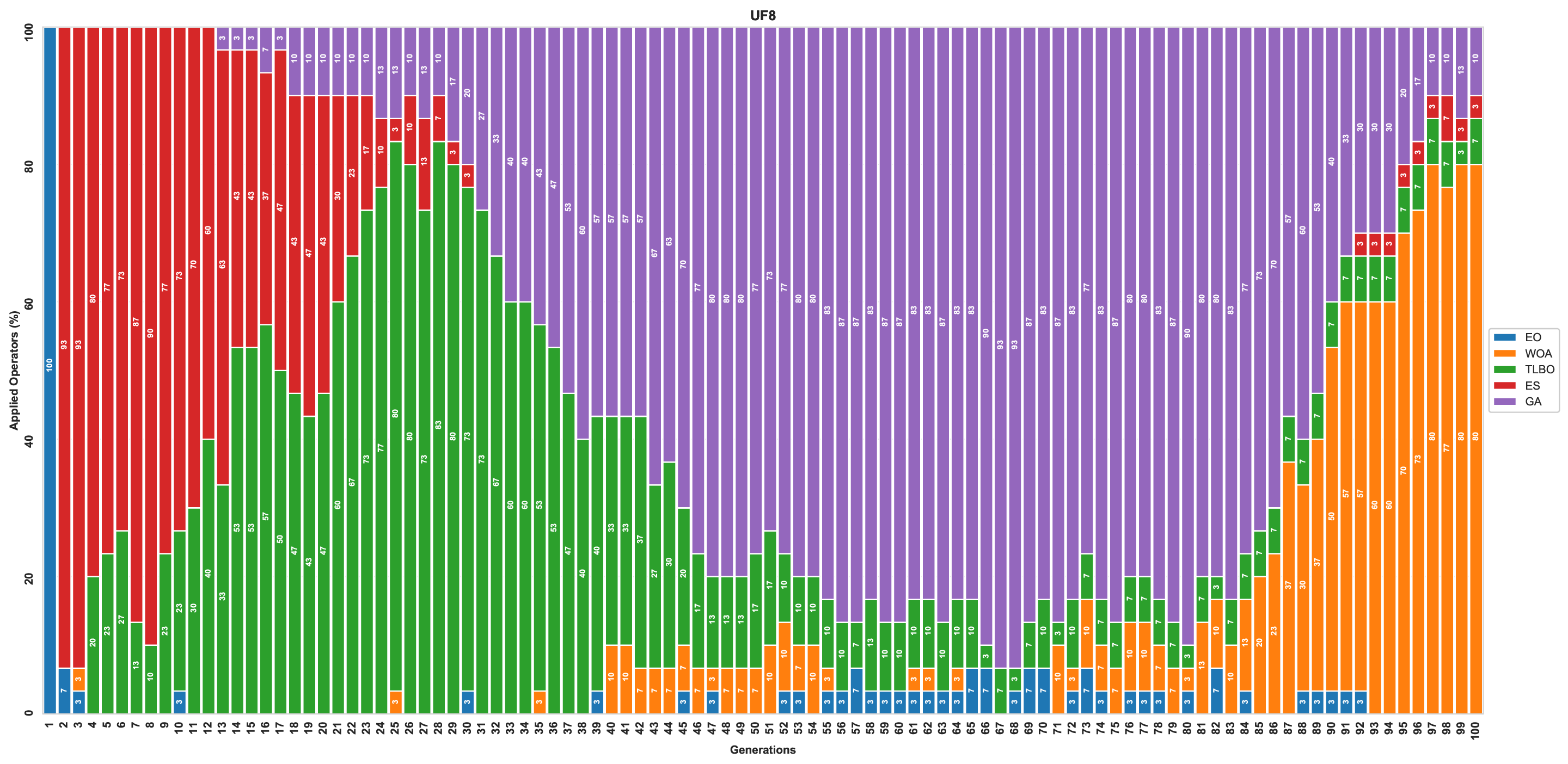}
    \caption{The percentage of each EA selected by the RL agent in each generation for the UF8 test function.}
    \label{fig.UF8Algplot}
  \end{center}
\end{figure}

\begin{figure}[H]
  \begin{center}
    \includegraphics[scale = 0.19, clip=true, trim=0.1cm 0.3cm 0.1cm 0.3cm]{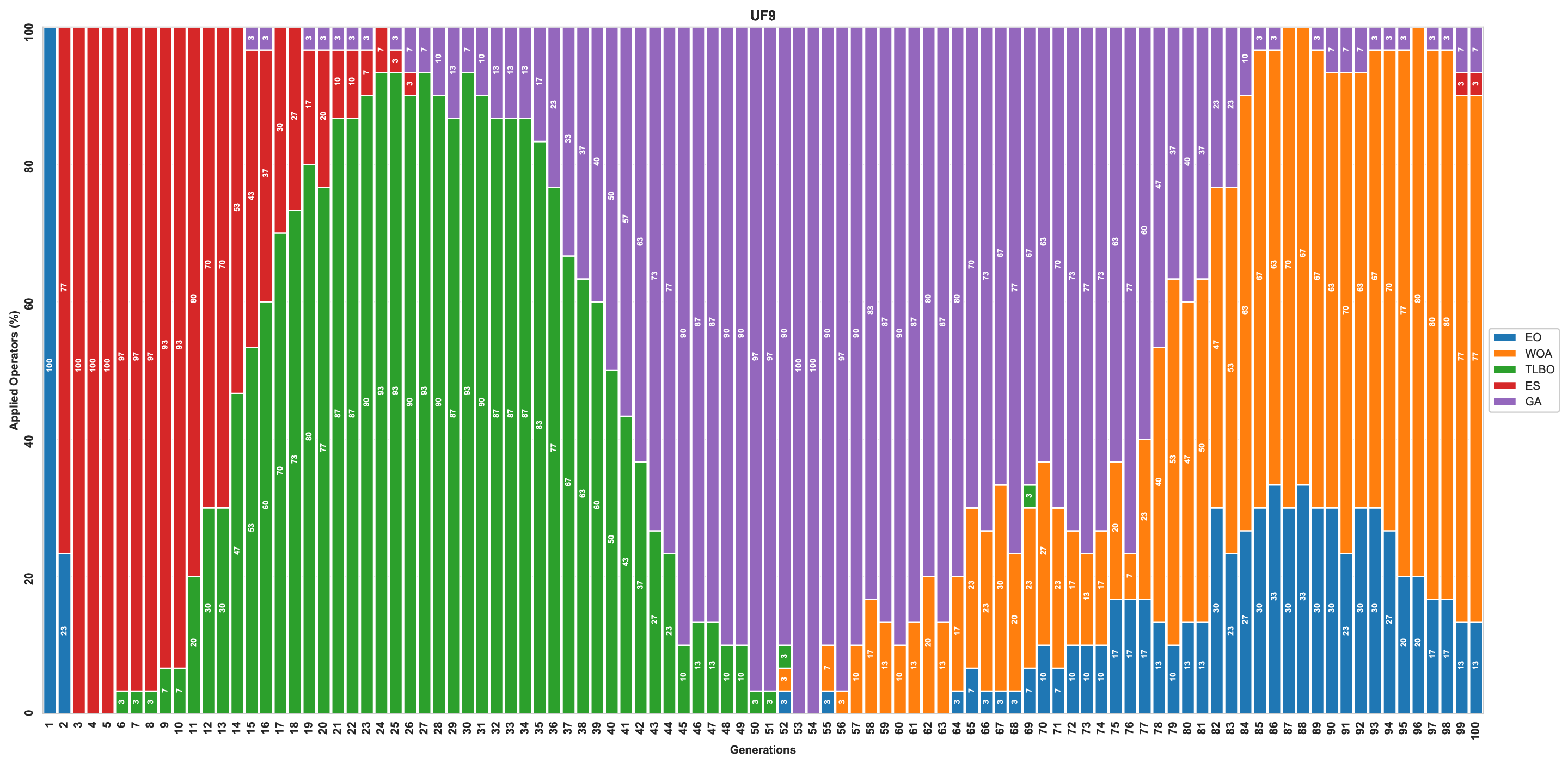}
    \caption{The percentage of each EA selected by the RL agent in each generation for the UF9 test function.}
    \label{fig.UF9Algplot}
  \end{center}
\end{figure}

\begin{table}[ht]
  \caption{The percentage of EA operators selected across different benchmark problems.}
  \label{tab.operatorcontribution}
  \center
  \scalebox{0.72}{
  \begin{tabular}{|c|c|c|c|c|c|}
    \hline
    Test Problem & EO Operator & WOA Operator & TLBO Operator & ES Operator & GA Operator\\
    \hline
    UF1 & 20.87\% & 1.10\% & 21.47\% & \cellcolor[gray]{0.7} 32.23\% & 24.33\% \\
    \hline
    UF2 & 22.87\% & 0.00\% & 1.17\% & 0.00\% & \cellcolor[gray]{0.7} 75.97\% \\
    \hline
    UF3 & 5.70\% & \cellcolor[gray]{0.7} 66.53\% & 3.40\% & 9.67\% & 14.70\% \\
    \hline
    UF4 & 7.20\% & 0.70\% & 21.73\% & 0.13\% & \cellcolor[gray]{0.7} 70.23\% \\
    \hline
    UF5 & 18.60\% & 11.90\% & 14.77\% & \cellcolor[gray]{0.7} 27.50\% & 27.23\% \\
    \hline
    UF6 & 0.97\% & 16.97\% & 17.17\% & 21.30\% & \cellcolor[gray]{0.7} 43.60\% \\
    \hline
    UF7 & 14.73\% & 0.27\% & \cellcolor[gray]{0.7} 36.77\% & 17.90\% & 30.33\% \\
    \hline
    UF8 & 2.53\% & 11.47\% & 24.63\% & 13.93\% & \cellcolor[gray]{0.7} 47.43\% \\
    \hline
    UF9 & 7.80\% & 18.83\% & 24.17\% & 13.47\% & \cellcolor[gray]{0.7} 35.73\% \\
    \hline
    UF10  & \cellcolor[gray]{0.7} 34.97\% & 0.07\% & 6.47\% & 24.00\% & 34.50\% \\
    \hline
    \hline
    Average & 13.62\% & 12.78\% & 17.17\% & 16.01\% & \cellcolor[gray]{0.7} 40.41\% \\
    \hline
  \end{tabular}}
\end{table}

\section{Discussion}\label{Sec.Disc}
Selecting and optimizing a specific evolutionary algorithm for a particular optimization problem is a difficult task. To address this challenge, this paper presents an adaptive MOEA (R2-RLMOEA) where a reinforcement learning agent guides the search process. We evaluated the performance of the R2-RLMOEA algorithm across the CEC09 benchmark problems UF1-UF10. These benchmark problems cover a range of test scenarios with two and three objectives. We systematically and statistically compared the performance of the R2-RLMOEA algorithm against five other R2-based MOEAs and a random operator selection of MOEA. This comparison provided valuable insights into how the RL-based agent selects various MOEAs operators during the optimization process, a crucial aspect of its design. We critically assessed the algorithm’s performance to identify its strengths and weaknesses. We used the IGD and SP metrics to evaluate how well R2-RLMOEA performed. These metrics are commonly used to measure convergence and diversity and assessed R2-RLMOEA compared to other algorithms.

Even though the R2-RLMOEA performs the best on average across all datasets and thus has been demonstrated to be the most versatile of all algorithms compared in this study, the results show that there is scope for improvement in the R2-RLMOEA algorithm, based on the observation that MOMBI-II (UF1), R2-ES and R2-TLBO (UF4) and MOMBI-II and R2-EO (UF5) performed best based on SP and MOMBI-II (UF6), R2-TLBO (UF3) and Random Opt algorithm (UF1) performed best based on the IGD indicator. Improvements in R2-RLMOEA would focus around additional optimisation of specific EAs, perhaps broadening hyperparameters ranges, to address challenges associated with different datasets/environments and/or incorporating additional EAs that are known to be better suited to specific environments/data in the R2-RLMOEA algorithm.

The findings of the study highlight the potential of RL in optimizing algorithm selection. The RL-based agent exhibited an advanced capability to navigate through various MOEAs and made informed choices that improved the optimization process. Our analysis of EA selection showed that certain EAs perform better during early-stage optimization generations, while others show good performance during the final stages. After studying the results of the RL agent in the successful benchmarks, we observed and provided evidence that the ES algorithm has strong exploration capabilities and is best suited to the initial stages of optimization. On the other hand, the GA and TLBO algorithms maintain a good balance between exploration and exploitation. Finally, the EO and WOA are mostly been used during the last generation, revealing their exploitation features. This adaptability is particularly important in multi-objective optimization tasks, where a one-size-fits-all approach is usually ineffective. 

In the RL component of R2-RLMOEA, we employed the DDQN model, which is a model-free and off-policy method. DDQN involves experience replay, which makes it highly stable and effective in various environments. However, one disadvantage of using DDQN is the costly training network time compared to EA optimization. Despite this, off-policy methods enable R2-RLMOEA to be well-prepared before the exact optimization process (by training the network early to be better prepared when encountering any state in the future). During optimization, there is no difference between using an agent-based RL and an EA without an RL agent. After analyzing the results of various tests, it can be concluded that the R2-RLMOEA algorithm is a highly effective tool for tackling complex optimization challenges over average performance across all benchmarks. Its dynamic approach to selecting MOEAs is particularly noteworthy and highlights the benefits of incorporating reinforcement learning into EAs. These findings not only showcase the potential of this exciting field of research but also pave the way for future advancements.

\section{Conclusion and future work}\label{Sec.Conclusion}
Our work introduces the R2-RLMOEA, an adaptive algorithm that combines multiple optimization operators for efficient multi-objective optimization. Our algorithm features a DDQ network that selects the appropriate EA based on the process condition. We used five EAs (GA, ES, TLBO, WOA, and EO) as well as an MOEA with randomly selected operators.  These are chosen adaptively to improve the optimization process. The R2 indicator is utilized to convert each SOEA into an MOEA and construct an RL reward architecture. The algorithm was evaluated using the IGD (for convergence and distribution) and SP (for distribution) criteria on multi-objective CEC09 benchmarks, with results demonstrating that our agent-guided MOEA (R2-RLMOEA) significantly outperforms all other algorithms on average across 10 benchmarks ($p<0.001$) and is significantly better than all algorithms for specific benchmarks UF8-UF10, as the three-objective benchmarks.

For future work, our plan is to incorporate advanced multi-objective frameworks, both non-indicator-based and indicators, to enhance the conversion of SOEAs to MOEAs. Additionally, we will explore the utilization of more EAs to enhance the convergence and distribution of our solutions across all conceivable test functions during the optimization procedure. We also plan to implement advanced RL algorithms, particularly policy-based methods that are suitable for optimizing continuous variables and parameters.
\bibliography{bibliography}
\end{document}